# From Feature-Based Models to Generative AI: Validity Evidence for Constructed Response Scoring

Jodi M. Casabianca[1*], Daniel F. McCaffrey[2], Matthew S. Johnson[2], Naim Alper[2], and Vladimir Zubenko[2]

[1]BroadMetrics, Edison, NJ USA

[2]ETS, Princeton, NJ USA

[*]Corresponding author: Jodi M. Casabianca, jodi@broadmetrics.co

**Author Note**

Jodi M. Casabianca  https://orcid.org/0000-0002-1644-6731

Daniel F. McCaffrey  https://orcid.org/0000-0003-1196-5273

Matthew S. Johnson  https://orcid.org/0000-0003-3157-4165

## Abstract

The rapid advancements in large language models and generative artificial intelligence (AI) capabilities are making their broad application in the high-stakes testing context more likely. Use of generative AI in the scoring of constructed responses is particularly appealing because it reduces the effort required for handcrafting features in traditional AI scoring and might even outperform those methods. The purpose of this paper is to highlight the differences in the feature-based and generative AI applications in constructed response scoring systems and propose a set of best practices for the collection of validity evidence to support the use and interpretation of constructed response scores from scoring systems using generative AI. We compare the validity evidence needed in scoring systems using human ratings, feature-based natural language processing AI scoring engines, and generative AI. The evidence needed in the generative AI context is more extensive than in the feature-based scoring context because of the lack of transparency and other concerns unique to generative AI such as consistency. Constructed response score data from a large corpus of independent argumentative essays written by 6-12th grade students demonstrate the collection of validity evidence for different types of scoring systems and highlight the numerous complexities and considerations when making a validity argument for these scores.

## 1. Introduction

Natural language processing (NLP) solutions for automatically scoring constructed responses (CR) are well established and used broadly in standardized testing for written, spoken, and short answer responses (Attali & Burstein, 2006; Chen et al., 2018; Dikli, 2006; Leacock & Chodorow, 2003; Shermis & Burstein, 2003; Xi et al., 2008; Zechner et al., 2009). In many applications, a set of features is selected that is intended to represent the construct as defined by the scoring rubric and combined to predict human ratings. These features are handcrafted and trained by NLP scientists to be extracted from the response and then used in a model to generate an overall score for the response. Some new artificial intelligence (AI) solutions, specifically approaches using generative AI such as GPT-4o, are not engineered to produce features based on the same principles of NLP to match the scoring rubric and construct. Instead, generative

AI approaches involve prompting an underlying large language model (LLM) to produce outputs such as a rating, sometimes with relatively little training or fine-tuning of the LLM to the specific task of scoring. It is difficult to explain how these generative AI approaches obtained their outputs since the LLM has billions of parameters, but they do offer capabilities that were not previously available without extensive effort by experts.

While the future of assessment will undoubtedly make extensive use of AI and given the accessibility of LLMs, it is important to place guardrails around their use in educational assessment so that they are used responsibly. Discussions on ethical AI and standards for using AI in education are prevalent in the literature (Bulut et al., 2024; International Test Commission & Association of Test Publishers, 2022; Johnson, 2024). This paper uses writing response data exploring the utility of LLMs in the context of CR scoring to highlight how validity arguments might be specially structured when using generative AI. The following sections discuss established validity frameworks for evaluating human ratings and automated scores from traditional CR scoring systems. We then introduce an additional set of validity evidence that should be collected and documented when using generative AI in CR scoring systems. We demonstrate the curation of validity evidence using empirical data from a persuasive essay task, scored with humans, an automated scoring engine, and multiple LLMs. We conclude with suggestions for practitioners.

## 2. Validity Framework for CR Scoring

The *Standards for Educational and Psychological Testing* (American Educational Research Association [AERA], American Psychological Association [APA], & National Council on Measurement in Education [NCME], 2014) provide high level guidelines for demonstrating the validity (including fairness) of CR scoring. Even though the document was written when the use of NLP and AI for scoring was just emerging into the mainstream, the *Standards* made occasional references to automated scoring. Its guidelines can serve as the basis for evaluating validity for scoring constructed responses with generative AI. Other publications such as Bennett and Zhang (2015) and Williamson et al. (2012) elaborated approaches for collecting validity evidence for constructed response scores from feature-based AI scores. To provide more detailed guidance on collecting evidence for AI scores, ETS published its *Best Practices for Constructed-Response Scoring* (McCaffrey et al., 2022) which included a framework for establishing validity evidence for CR scores from human raters or AI. We build on these sources to create a framework for generative AI scoring, first reviewing the frameworks for human ratings and feature-based AI scoring and then extending those guidelines to generative AI but framed around how generative AI is distinctly different from feature-based scoring.

### 2.1 Evidence for Human Ratings

The validity evidence for CR scores, of any kind, starts with the task design. Ideally task and test design are conducted within a formal framework such as evidence-centered design (ECD; Mislevy et al., 2003). Part of task design for CR items is the scoring rubric for judging the responses and assigning scores. When humans provide the scores, the key to the validity of the scores is the ability of the human raters to consistently use the rubric as intended. To ensure this occurs, there are several processes involved with managing human rating. Decisions made throughout the design of the scoring system should be made with principles of validity, reliability, and fairness, and they should be documented as part of the curation of validity evidence. For example, the training materials developed for raters should include additional details on the task including annotated exemplars to ensure that raters will apply the rubric as it was intended. Raters should be trained and then evaluated before scoring to ensure they are qualified to score. To the extent possible, raters should be recruited from a diverse population with a specified skill set qualifying them to rate responses for the assessment. During the rating process, raters should be monitored to make sure that they are consistent and accurate. This means that during the design of the system before scoring occurs, decisions must be made on how to collect data to measure consistency and accuracy.

Table 1 provides a list of five different types of evidence that should be collected to make a validity argument for scores (as per the 2014 *Standards*) and examples of specific pieces of evidence that should be documented and/or collected to make the argument for scores based on human ratings. For example, evidence of internal structure might include a documented link between the item or task and the construct definition. For a CR item, there should also be a documented link between the scoring rubric and the construct definition. This documentation may include a memorandum or report summarizing how the task allows the test taker to demonstrate knowledge, skills and abilities related to certain aspects of the construct. Other traditional evidence, such as factor analyses, are included here. Importantly, we should demonstrate that the human rating process minimizes construct-irrelevant variance and therefore a system for monitoring the interrater agreement and accuracy should be planned, implemented, and documented.

Other types of evidence have been heavily discussed in the literature and so have the methods used to detect unfairness, which are included in the last set of rows of Table 1 as a type of evidence to create a validity argument. For example, validity coefficients (e.g., correlations) from predictive validity studies (as appropriate), inter-item correlations, and content reviews are standard validity evidence collected for selected responses and CR items and their approaches are well-established. However, the human rating component introduces added opportunities for construct irrelevant variance. In CR scores, for example, we want to expand the traditional sense of response processes by adding the response process of the rater using the rubric.

## 2.2 Evidence for Automated Scores from Models Predicting Human Ratings

Johnson and Zhang (2024) discuss the different approaches to automated scoring, ranging from using handcrafted features to prompting generative AI models, and the tradeoff between interpretability and accuracy when advancing from highly interpretable models to models with low interpretability. Here we distinguish models predicting human ratings from generative AI because when prompting generative AI, there is no use of human scores in developing the AI models, unless the LLM is fine-tuned. Otherwise, there is no explicit connection to human ratings.

### 2.2.1 Construct Feature-based Models

The traditional use of AI in CR scoring relied on experts such as NLP scientists and linguists to create hand-crafted features quantifying different components of written text or spoken responses (Shermis & Burstein, 2013). The NLP scientists' expertise in creating and purposefully combining features to provide construct coverage and alignment with the scoring rubric ensured a human-in-the-loop approach to AI scoring. The statistical models once used to combine features were simple, for instance, multiple linear regression.

For construct feature-based AI scores, there is a chain of evidence that goes from the automated scores back to the human ratings due to how the engine is trained. Thus, it is important to make a validity argument for the human ratings and then the machine scores. Ideally, tasks that are intended to be automatically scored are designed with automated scoring in mind, and the procedures and decision-making should also be documented. If human ratings are then used to train automated scoring models, we prefer to start with "high quality" ratings based on sound practices as described in the previous section and in Table 1 and with at least a satisfactory psychometric profile.

**Table 1** Validity evidence for different types of CR scoring systems

| Validity Evidence | Human Ratings | Construct Feature-based AI Scores | General Linguistic Feature-based & Embeddings-based AI Scores | Gen AI Scores |
|---|---|---|---|---|
| **Internal Structure** | | | | |
| • Link between item and construct definition | ✓ | | | |
| • Link between scoring rubric and construct | ✓ | ✓ | | |
| • Features/rubric aligned with construct; no construct-irrelevant features | ✓ | ✓ | | |
| • Features trained on representative sample (*pre-training or fine-tuning data) | | ✓ | ✓ | ✓* |
| • Chain-of-thought results show consistency with construct definition and aspects of the rubric | | | | ✓ |
| **Relations to External Variables** | | | | |
| • Correlation with section/total scores | ✓ | ✓ | ✓ | ✓ |
| • Correlation with other tests/external variables | ✓ | ✓ | ✓ | ✓ |
| • Convergent/discriminant validity studies | ✓ | ✓ | ✓ | ✓ |
| • Correlation of features with other variables | | ✓ | | |
| • Contrasted group/predictive/concurrent studies | ✓ | ✓ | ✓ | ✓ |

**Response Processes**

| | | | | |
|---|:---:|:---:|:---:|:---:|
| • Review of response processes (item design) | ✓ | | | |
| • Review of response processes for raters via think-alouds, rater feedback/surveys | ✓ | ✓ | ✓ | ✓ |
| • Expert review of responses and scores | | ✓ | ✓ | ✓ |
| • Review of scoring of atypical responses | | ✓ | ✓ | ✓ |
| • Evaluation of chain-of-thought outputs & comparison to expert annotation | | | | ✓ |

**Test Content**

| | | | | |
|---|:---:|:---:|:---:|:---:|
| • Expert review of item/rubric/features for content coverage and relevance | ✓ | ✓ | | |
| • Evaluation of feature value variation | | ✓ | | |
| • Inter-item and item-section correlations | ✓ | ✓ | ✓ | ✓ |
| • Expert review of scores/responses and annotations for score levels | | ✓ | ✓ | ✓ |
| • Training materials, exemplars, calibration test, certification process, etc. designed to be aligned with construct (and have no construct irrelevant features) | ✓ | | | |
| • Documentation of model/prompt/feature selection | ✓[1] | ✓ | ✓ | ✓ |
| • Selection rationale for LLM (if applicable) | | | | ✓ |
| • Evaluation of rater accuracy/interrater reliability | ✓ | | | |
| • Concordance with human scores (initially and ongoing) | | ✓ | ✓ | ✓ |
| • Reproducibility of model scores: Studies showing reproducibility and consistency of LLM scores over time. Documentation of the variability of LLM scores and how that affects reported score reliability. | | | | ✓ |

**Consequences of Use**

| | | | | |
|---|:---:|:---:|:---:|:---:|
| • Analysis of unintended consequences | ✓ | ✓ | ✓ | ✓ |
| • Analysis of intended consequences | ✓ | ✓ | ✓ | ✓ |

**Fairness**

| | | | | |
|---|:---:|:---:|:---:|:---:|
| • Item fairness reviews | ✓ | | | |
| • Differential item functioning analysis | ✓ | ✓ | ✓ | ✓ |
| • SME review of saliency bias | | ✓ | ✓ | ✓ |
| • Human-human agreement by subgroup | ✓ | ✓[2] | ✓[2] | ✓[2] |
| • Human-machine score comparisons | | ✓ | ✓ | ✓ |
| • Differential feature functioning | | ✓ | | |
| • Saliency methods for subgroup differences | | | ✓ | ✓ |
| • Differential algorithmic bias analysis | | ✓ | ✓ | ✓ |
| • Fairness metrics for LLMs | | | | ✓ |
| • Report on LLM pretraining data | | | ✓ | ✓ |

1. Includes documentation of rater selection (expertise, experience, training).

2. For AI systems, this applies to the fairness of the human ratings used to train/fine-tune the model.

The traditional NLP feature-based automated scoring approach requires similar evidence to the human ratings with some differences due to the nature of the scoring process. While human raters apply a scoring rubric and make judgments, a scoring engine extracts information from the responses to reflect different construct-relevant features. A statistical prediction model is then trained to predict the human rating using those features. In an engine that evaluates writing ability, features may include grammar, usage, mechanics, style, and organization (Attali, 2007; Attali & Burstein, 2006). In an engine that evaluates spoken responses, features might include words per minute, average pause length and others for accuracy and pronunciation (Xi et al., 2008). The set of features should be combined to represent the construct. Validity evidence may include documented links between the feature set and the construct definition. It may also include a summary of the prediction model weights to determine the extent to which the combination of features and weights correspond to how raters should be combining information about the response to derive the score.

For evidence of internal structure of the test (AERA, APA, & NCME, 2014, p. 16), we would collect evidence analogous to what we would collect for the human ratings—documented links between the features and scoring rubric; we might perform factor analyses or inter-item correlations with the machine scores and other item scores on the test; and we also monitor the AI scores by comparing them to human ratings of the same responses. This concordance should be established during model evaluation, but also during operational scoring to ensure there are no issues with predictions on a new sample of test takers. In addition, we should train the features on data not used for model-building or evaluation and the sample should be representative of the target test taker population.

Much of the other types of evidence for AI scores would be collected at the time of the initial model evaluation. A model-level evaluation includes an examination of concordance between the human ratings and the machine scores. Metrics might include standardized mean differences, correlations, QWK, etc. (Williamson et al., 2012). Recent arguments have been made to include a measure of the accuracy of predicting the human true score via the proportional reduction in mean square error (PRMSE; McCaffrey et al., 2025).

An impact-level evaluation might compare the CR section scores (based on machine scores) to other section scores, or other completely external scores if available. We also want to compare section and total test scores based on the machine scores to corresponding section and total scores based on human ratings to understand the size of the differences at that level. For evidence that the response processes are appropriate, we rely on annotations by subject matter experts of a selection of responses to make sure that there is justification for the scores given by the engine. This type of qualitative analysis should be performed at all score levels. In addition, we also must provide evidence that the engine is properly handling "atypical" responses which may be in the wrong language, off-topic, a copy of the prompt text, etc. The scoring engine must be trained to either flag these responses so that they can be hand-scored or trained to assign an appropriate score to them (likely a 0). Analyzing the effectiveness of any "advisories" or flags the engine might use is important to ensure that these responses are detected and scored appropriately. This minimizes the chance that atypical responses that deserve a score of 0, for example, do not get some higher score.

Checking for fairness is of vital importance, however simply checking for fairness is insufficient. Fairness should be part of the *design* of the scoring system. To start, in addition to the fairness checks that should occur on the human ratings, we need to ensure that the samples used to train features, build the models, and evaluate the models are large and representative of the test taker population. This is important because there might be differential response styles by subgroup and the engine must be trained to score those appropriately. In addition, if there is an imbalance in the composition of the training sample and many groups constitute only a small proportion, there may be inadequate representation. Oversampling or weighting up small groups in the training sample may be a solution to this (Zhang, 2013). An evaluation of the demographic composition of the sample and how it might impact fairness should be documented.

Fairness checks during model evaluation often involve a comparison of human and machine score means, by subgroup, via a standardized mean difference (Williamson et al., 2012). In addition, comparing the QWK by subgroup may also be helpful to understand if agreement is degraded for different test taker groups. The *Best Practices* (McCaffrey et al., 2022) discuss challenges with subgroup analyses, including that the small sample sizes for groups would prevent these groups from being assessed for fairness. Recent work utilizes empirical Bayesian methods to better estimate SMD and flag potential fairness issues in the small sample case (Kwon et al., 2025). In addition to basic SMDs and QWKs, we might also run differential item functioning (DIF) analyses—first for the human scores and then for the machine scores, to compare the DIF results for humans and machines. Differential feature functioning (Zhang et al., 2017) and differential algorithmic functioning (Suk & Han, 2024) are also appropriate analyses to better understand the extent to which the engine features or the model may be disadvantaging certain groups. We may perform all of these together to collect evidence on the fairness of scores.

### 2.2.2 Models Using General Linguistic Features and LLM Embeddings

For tests of content knowledge or reading comprehension, models based on general linguistic features are often used for scoring constructed responses because construct-based features are unavailable or are too item specific. In these models, features of the text such as n-grams or word-grams are used to predict human ratings to create the scoring model (Leacock & Chodorow, 2003). More recently, embeddings from language models have served as the general linguistic features for developing the scoring models (Mayfield & Black, 2020; Ormerod et al., 2021; Riordan et al., 2017; Rodriguez et al., 2019). These models are now used more broadly for tests of content knowledge or comprehension.

A construct feature-based model, general linguistic feature-based model, and a model based on LLM embeddings all share the same "chain of evidence" link via the prediction of human ratings. This link somewhat mitigates the concern of reduced transparency for the larger models using general linguistic features or embeddings, even though there might still be many thousands of parameters and an increased risk of the predictions relying on construct irrelevant features of the response.

Most of the evidence required for these models will be similar to the construct feature-based models but not all will be available. For example, it may be infeasible or inappropriate to match engine features to rubric indicators. In content engines such as ETS' c-rater (Leacock & Chodorow, 2003), often a written response is evaluated using many generic linguistic features and/or keyword indicators to capture specific content in the response or specific written structures not explicitly related to the content. The responses may be parsed into n-grams which could generate thousands of "features". In this case, the features are not necessarily understandable, and the machine learning models used for prediction are more sophisticated than traditional regression models meaning the "weights" are not something that can be easily examined or immediately understood. The same would be true if using embeddings from an LLM. As a result, the transparency of the automated scoring process is reduced and the validity evidence for the machine scores relies more heavily on the quality of the human ratings used to train the models (McCaffrey et al., 2022). We might also require a very strong link between the scoring rubric and the construct definition and demonstrated high agreement between the human rating and AI scores during model evaluation and monitoring. So while we may collect similar evidence for this branch of models, we weight certain evidence (related to the human ratings) heavier.

Some of the evidence we propose for generative AI scores will be useful for making a validity argument for these scores as well. Table 1 reflects these differences. For example, saliency methods to understand the importance of different features or embeddings may provide evidence that the scores resulting from these models lead to meaningful interpretations or that they do not contain construct irrelevant features.

### 3. Validity Evidence for CR Scoring Systems Using Generative AI

Generative AI is distinctly different due to the nature of the approach in generating scores. In this case, because the "engine" is generating an output in response to a prompt, there is no principled or explicit system of deriving a model or selecting features specific to the scoring task, and the score is not based on a prediction of a human rating. In addition, LLMs are trained on unknown data and scores from prompting LLM are not fully reproducible – repeated scoring of the same response can produce inconsistent results. As such, the types of validity evidence we might collect for feature-based AI scoring should be adapted for generative AI applications. Figure 1 organizes the various concerns with using generative AI for scoring.

### 3.1 Evidence to Address Issues Related to Unknown Training Data

The data used by LLM developers for pre-training is typically a collection of corpora and scrapes of text from the internet. What is in this data and how it impacts the LLM's completions on the task is unknown. There are two main tasks to incorporate into the collection of validity evidence to demonstrate that the associated risk with this issue is minimized: defensible selection of LLM based on data sources and a careful fairness analysis and/or fairness mitigation plan.

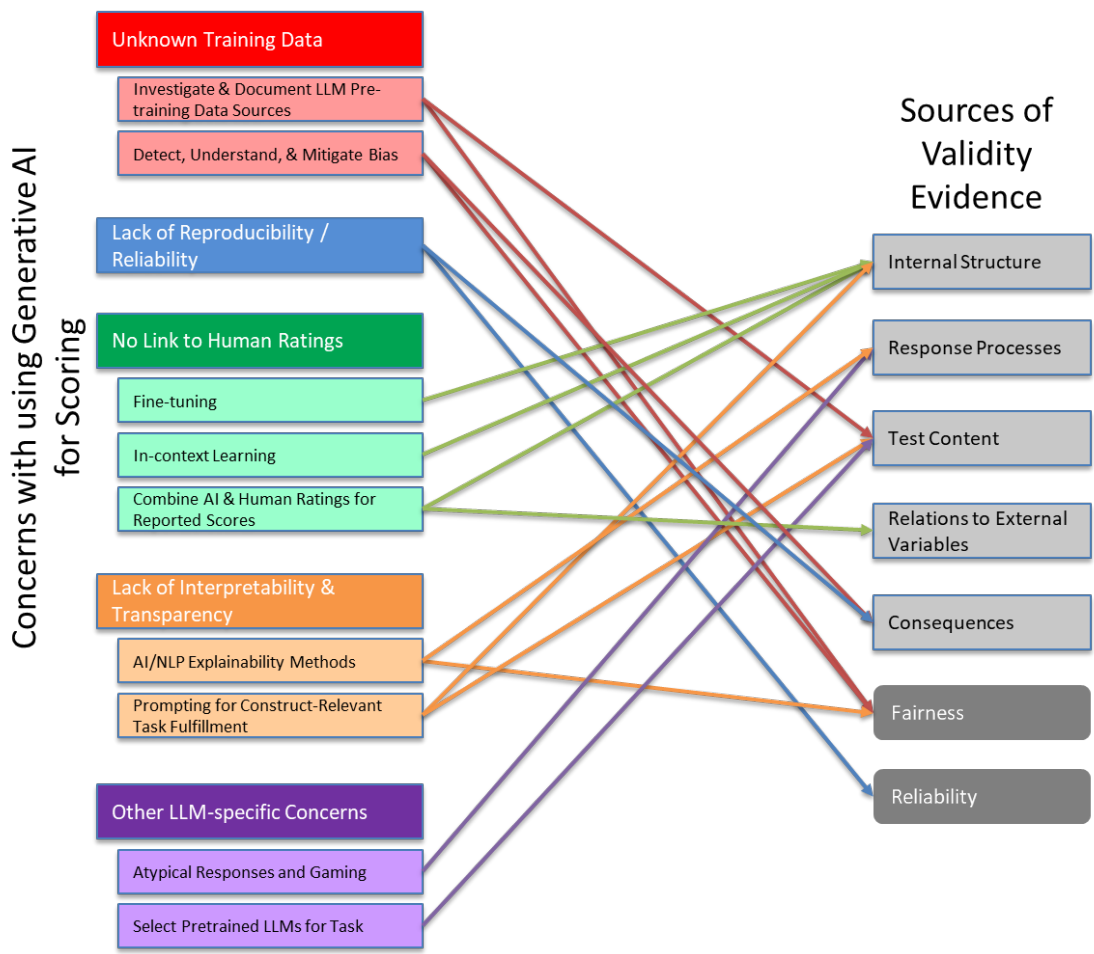

**Fig 1** Connection between the concerns with using generative AI for scoring constructed response tasks, the different actions we take to address them, and how they relate to different sources of validity evidence to support the use and interpretation of the scores.

### 3.1.1 Investigate and Document LLM Pre-training Data Sources

There are several different LLMs available, and each may have different versions from which to choose. We posit that the LLM should be carefully researched before selection. The primary concern is the data used for pretraining. While this is not disclosed by most developers, some do provide descriptions of their data sources in their technical reports. For example, Meta discloses that LLaMA was pre-trained using the following corpora: English CommonCrawl, C4, Github, Wikipedia, Gutenberg and Books3, ArXiv, Stack Exchange, and Tokenizer (Touvron et al., 2023). Qwen's technical report has a section on "pre-training data" which states: "All Qwen3 models are trained on a large and diverse dataset consisting of 119 languages and dialects, with a total of 36 trillion tokens" (Yan et al., 2025). This type of investigation is important because we want to ensure that the model is trained on data that are relevant to the target population. We also want to ensure that there are no explicit biases in the data and if there was any "de-biasing" conducted during pre-training. We cannot be sure of this, but we can consider our target population and the source of the texts and see if there is a gross mismatch. For example, suppose we want to use an LLM to score essays written in Chinese. Though there is research to show that large proprietary models like GPT4 and Claude Sonnet can accurately score Chinese language essays (Gao et al., 2025), are they really the *best* LLM for the task? Would a Chinese LLM like Qwen (Yang et al., 2025) or DeepSeek (DeepSeek AI, 2024) be better suited since they are likely based on the scraping of different websites or used Chinese corpora? We may not be able to definitively answer these questions, but the burden is on the data scientist, engineer, or the person selecting the LLM to do the due diligence to understand the sources of the data used to pretrain the LLM and conduct experiments. The decision-making should be documented as well as an explicit statement about any investigations conducted on the pre-training data.

### 3.1.2 Design a System to Detect, Understand, and Mitigate Bias

We must be diligent with fairness checks when using an LLM for scoring due to the biases that may exist in the pre-training data. In traditional AI scoring in which we collect a sample of human rated response data to train and tune the model, we can theorize the types of biases that might manifest. In English language testing of non-native speakers, certain culturally-specific response styles can interact with the features of the AI model that may penalize or benefit them. For example, Japanese secondary school learners are trained in short essay writing (Reichelt, 2009; Rinnert & Kobayashi, 2009), but we have observed human rating patterns that heavily depend on response length (giving shorter essays lower scores). AI scoring models trained on these human ratings (such as e-rater) can exaggerate this dependency (Chodorow & Burstein, 2004). As a result, in these cases, we might see AI scores for Japanese test takers being lower than human ratings. Basic fairness analyses such as standardized mean differences (SMD) comparing human and AI scores might catch score differences from such patterns. To examine fairness from different perspectives, we may also perform additional fairness analyses based on a comprehensive set of definitions of fairness as described in recent literature (Johnson et al., 2022; Johnson & McCaffrey, 2023). DIF analyses on the LLM scores using methods used for DIF analysis for CRs scored by humans (Moses et al., 2013) is another way to test for unfairness.

Simply examining SMDs is not sufficient in the LLM scoring scenario, because the investigation conducted as a result of the large SMD will not get to the root cause. We will be unable to use features of the engine to understand the nuances or interactions, or even theorize why there might be subgroup bias. Approaches to explainability might be helpful to understand bias, but if using an off-the-shelf model, we are limited to manipulating inputs and outputs. For example, we might run experiments where responses are perturbed to resemble or reveal a culturally-specific or demographic-group-specific feature (dialects, pronouns, etc.), to then observe how the LLM's output changes. We may also use investigative prompting to understand biases. For instance, Kaneko et al. (2024) show that using chain-of-thought (CoT) prompting to make LLMs explicitly reason about gendered terms reduces biased misclassifications, highlighting CoT as a practical tool for detecting and mitigating social bias in model outputs. Johnson and Zhang (2024) performed an experiment to determine if GPT-4o differentially scored responses based on their own identified racial/ethnic groupings of test takers and found that it can better predict race/ethnicity than the scores. These guesses had an impact on GPT-4o's scoring, even when adjustments were made for real differences in human scores. This indicates that GPT-4o might be considering information related to race/ethnicity from essays. It has been shown that LLMs can infer native language as well (Zhang & Salle, 2023). This type of deeper analysis aimed at improving transparency/explainability might help to establish if there is a relationship between the LLM-identified subgroup and the scores, in a way allowing the LLM to identify its own bias. Another approach to explaining bias would be to analyze NLP features to understand what parts of the response are differentially correlated to the LLM scores, though this is only possible in cases for which NLP features are available.

Ideally, we would actively attempt to mitigate bias. To our knowledge, we have seen very few CR scoring specific discussions of LLM de-biasing in the literature (Kwako, 2023; Kwako & Ormerod, 2024), but some of these strategies might be suitable for a high-stakes assessment program that want to use LLM for operational scoring. Gallegos and colleagues (2023) provide a comprehensive taxonomy of techniques for bias mitigation in LLMs. Intra-processing and post-processing mitigation strategies can be implemented with an off-the-shelf LLM. For example, there are several procedures that involve modifying token probabilities to increase the diversity of the completions, or constraining text generation by blocking certain word- or n-grams that would be considered biased or offensive. However, since the text generation in the AI scoring context produces a number, the efforts to minimize bias in the scoring process may not be as straightforward as efforts to minimize bias in a text summary or translation. Because of the type of output, the ideal location in the pipeline to apply mitigation techniques is likely during pre-processing or in-training, where we can control how the LLM uses the training data so that it impacts how the task (of scoring) is performed. Further research is needed in this area to understand how it can be useful in CR scoring. Certainly, if fine tuning is conducted by the AI score developer, it should be conducted using samples with sufficiently large numbers of test takers from all relevant groups and methods to correct biases should be implemented.

These steps to minimize unfairness, which involve quantifying, understanding, and explaining bias, as well as mitigating it, should be an integral part of the pipeline that is continuously reviewed and refined.

## 3.2 Evidence to Address Issues Related to LLM Reproducibility/Reliability

The output from an LLM is not deterministic like a typical model prediction. For example, in a feature-based AI scoring model, we might use a basic regression model or even a machine learning model such as support vector regression to predict human scores. We estimate model parameters and use those with feature values to predict human scores for newly collected responses. Given the same set of feature values, we will always get the same prediction. Studies show that LLMs can have varying levels of consistency across runs at the same time and over time (Hackl et al., 2023; Tang et al., 2024). The GPT family of models has shown high consistency across runs in several studies (Hackl et al., 2023; Pack et al., 2024; Seßler et al., 2025).

There are multiple aspects of LLMs that prevent them from providing the same score on every occasion for the same response. First, there is a probabilistic component to the underpinnings of the LLM, which generates different outputs (i.e., scores). Specifications like the temperature or the approach to sampling the tokens can affect the variability in LLM outputs to the same prompt (Tang et al., 2024). To maximize consistency, we might set the temperature to be very low, or 0 (Si et al., 2023). If the temperature is above 1.0 we should explain that choice since it leads to more randomness.

Temperature is not the only factor. The empirical example discussed later in this manuscript using GPT-4o and Gemini 1.5 Pro shows variability in scores even with temperature = 0. Another factor leading to variation is the method by which the LLM is accessed. Application programming interfaces (API) are version controlled and have deprecation dates when they would no longer be available. Since consumers of the APIs do not have control over this process, these services would not be suitable for long term reliability and consistency. If they are used, a monitoring process will be needed to ensure that the scores remain consistent over time to ensure temporal stability. More control can be had if LLMs are self-hosted or deployed on local machines or in the cloud, but even these would benefit from continuous observation for model drift. Interestingly, in one study, closed-source LLMs (GPT3.5, GPT4, GPTo1) provided relatively more consistent scores than open-source LLMs (LLaMA 3 and Mixtral) (Seßler et al., 2025).

To ensure that there is exchangeability between scores from different times, we should perform consistency analyses which may involve generating scores for the same set of responses over a period of time. This is similar to checking for interrater consistency and test-retest reliability in the traditional psychometric model. Though the results may not be exactly the same, we might set a standard for the amount of variability in scores that is considered acceptable. We would also check for systematic changes over time such as the scores all tending to increase or decrease or be more or less concentrated in a subset of the scale. We need to explicitly describe variability due to lack of determinism in the model as an extra source of variability in the scores and document it so that users of the scores are aware. Alternatively, we might collect multiple scores from the same LLM and average them, similar to what some testing programs do to mitigate errors in human ratings (e.g., some subjects in the *Praxis®* Assessments use the mean of two human ratings). Analyses exploring intra-LLM consistency or alternative scoring methods to account for variability should ensure that no group of test takers is affected more by this type of inconsistency than overall.

During operational scoring, regular monitoring of the consistency of the LLM is advised. We might use the same set of responses during each scoring period as a check, to remove the conflation between changes in the population and changes in the LLM. In human CR scoring systems it is called trend-scoring (Tan et al., 2009). If a population shift is detected, the test-set of responses used in trend scoring studies should be updated so that it is once again representative of the population with respect to major demographic variables or the type/correctness of responses.

### 3.3 Evidence to Address Issues Related to No Link to Human Ratings

Integral to the validity argument for AI scores is the link from the AI scores to the human ratings, which is linked to the scoring rubric, which is linked to the construct definition. Without the humans, the link back to the construct relies on the AI model's ability to properly simulate the human scoring process via the rubric. Without this link, there are some places in the pipeline in which we can incorporate human data, as well as attempt to connect the LLM scores to the construct definition directly.

### 3.3.1 Fine-tuning

In the absence of human ratings in the pre-training of the model, fine-tuning with a sample of human-scored responses will be the closest way to simulate the link we have in the prediction models. Decisions surrounding the fine-tuning methods should be made with consideration of the validity framework. This is similar to the decisions we make about training for human raters in that we need to ensure and document that the materials selected for use in the human scoring process (for training, evaluating raters, etc.) are related to the construct and do not introduce any construct irrelevant variance or unfairness. Decisions on the sample of responses used for fine-tuning and how many responses are available and used should be made carefully. Care should be taken to ensure that the sample represents various different groups of test takers and responses and that the scores provided are correct according to subject matter experts. For example, if the test is for English language learners from various language groups, it is important to provide examples from as many groups as possible. Using a convenience sample of prompt-completion pairs from only native Spanish speakers, for instance, may lead to bias when using the trained LLM to score test takers from other language groups. Achieving a sample that is balanced with respect to relevant demographic subgroups via stratified sampling may improve fairness metrics (Zhang et al., 2024).

Fine-tuning can be performed over several rounds in an iterative approach. Fine-tuning can also be done on a single task (e.g., scoring essays) or multiple tasks (e.g., scoring essays and providing feedback to test taker). Multi-task fine tuning may be useful in improving the results for a single task by using different versions or wording of the prompt (e.g. scoring essays with rubric and scoring essays holistically based on rubric).[1] However, in order to perform multi-task fine-tuning many more example responses are necessary relative to single-task fine-tuning. Descriptions of how the examples used for fine-tuning (including any public datasets) are appropriate for the particular task, as well as the adequacy of the fine-tuning approach, should be documented as part of the validity evidence.

An important caveat to note is that while a fine-tuned LLM may perform better for the task than the pre-trained LLM (Latif & Zhai, 2024; Liu et al., 2025; Liu et al., 2025), there is a risk that changes in the test taker population may show degraded performance in a new population. Therefore, we should be careful when tuning the LLM that we are not "micro-tuning" to a population that will not be relevant in the future. This concern is similar to concerns with traditional AI scoring models, which is why we need to be cognizant of the samples used for training and evaluation as well as making sure there is a monitoring system in place to catch population changes.

### 3.3.2 In-context Learning

In-context learning (ICL) is another way to train the LLM, but a more accessible choice because very little data is required compared to full fine-tuning. We suggest conducting experiments with ICL to demonstrate the advantages of a one- or few-shot learning approach. The literature shows mixed results. In some applications for AI scoring, there may be no benefit to one- or few-shot approaches (Chamieh et al., 2024; Liu et al., 2025). Other research has shown improvements in model accuracy compared to zero-shot prompting (Stahl et al., 2024).

NLP features may also be introduced to the scoring process with LLMs. For example, suppose we have values for 10 features for essay responses that provide information on writing quality. We could tune the LLM with this information and provide instructions on how this should be used to generate a score. Hou et al. (2025) explored this approach with different language models and using different subsets of features and found that in some cases the addition of feature data in the prompt led to improved human-AI agreement. In theory this may lead to better construct coverage, however, it will be unknown how the LLM will actually use this information. Sensitivity analyses demonstrating that the LLM does utilize the features (e.g., showing meaningful score differences with and without features) may serve as evidence that it does contribute.

### 3.3.3 Combining AI Scores and Human Ratings for Computation of Reported Scores

It is possible that a scoring engine can be responsive to aspects of the construct definition not detected by the human raters so that a combination of the scores either by summing or averaging them (sometimes in a weighted average or using the best linear predictor approach) (Breyer et al., 2017) may be more informative than either score alone. Combining human and AI scores also may improve overall test reliability because both human ratings and LLMs contain random errors and a combination will reduce the contribution of those errors to the scores. Combining LLM scores and human ratings for overall reported scores is another approach to estimating scores with links to humans. In addition, if NLP features are available, we might also use NLP features by combining them with LLM scores directly using best linear predictor, provided there are human ratings available for a sufficient subset of responses (Yao et al., 2019a, b).

Following from this, since multiple feature-based automated scoring engines and/or multiple LLMs will provide different scores and for different reasons, we might wish to consider how these scores can be combined, with or without human ratings. The validity evidence needed to support the use of those score combinations should include an analysis comparing the score to the human-based score, and possibly different combinations of scores, as well as validity evidence based on the score combination (relations to external variables, consequences, etc.).

---

[1] Note that the reference to "task" here is the task the LLM is trained to perform. This could be scoring an essay or text response. In the case of automated scoring, the LLM might be trained to score a specific type of item (an argumentative essay) or multiple types of items (argumentative or persuasive essay). Multi-task fine-tuning could be used to train the LLM to score one type of item (argumentative) with just different approaches.

### 3.4 Evidence to Address Lack of Interpretability and Transparency

In traditional AI scoring, we can quantify relationships between scores and features of the response to understand and decompose the meaning of the scores. With generative AI, we cannot establish these relationships, which would provide evidence that the scores are appropriate for the correct (construct-relevant) reasons. We can address this using explainability analyses and strategic prompting, including chain-of-thought prompting, to both impose construct relevant task completion as well as understand the (reported) reasoning for the scores generated.

#### 3.4.1 AI and NLP Explainability Methods

Explainable AI methods have been applied in many contexts, but they are still not widely used in educational research or educational assessment. Open-source LLMs provide access to weights, embeddings, and intermediate activations, enabling a broader range of explainability methods. Saliency maps, which show the parts of the input that most influence the model's output, might be useful in a qualitative analysis to understand what aspects of a response lead to differences in scores from expert human raters, NLP scoring engines, and LLMs. They provide an accessible way to see which words or phrases most strongly influence scoring decisions and are used in educational assessment contexts (Hoffman et al., 2018; Li et al., 2016). Saliency maps may also help illuminate how in-context learning is influencing the scores. The maps might be built from attention weights (which tokens), input token importance (gradient-based methods), or log-probability sensitivity (examine output changes after perturbing a token). Another technique is probing models, where classifiers are trained on hidden representations to test whether features like grammar errors, coherence, or topicality are encoded (Alain & Bengio, 2017). Hybrid systems combine open LLM embeddings with interpretable models (e.g., linear regressions on trait dimensions) which can yield human-readable scoring rationales while retaining predictive accuracy (Molnar, 2022). More research is needed to understand how to derive meaningful explanations from these results (Hoffman et al., 2018, 2023). As open-source LLMs become more readily available, some of these methods may prove useful in providing evidence for internal structure and/or response processes.

Black-box or closed-source LLMs, where the model's weights and internal representations are inaccessible, require prompt-based and external methods to enhance explainability by imposing a reasoning structure and/or exploring the reported reasoning. One widely used strategy is rationale generation, where the model is prompted to accompany its score with natural-language justifications or rubric-aligned reasoning, functioning as a form of post-hoc self-explanation (Cheng et al., 2024). A qualitative analysis (by subject matter experts) of many of these "annotations" can help provide validity evidence for the LLM-based scores. Similarly, chain-of-thought prompting elicits step-by-step reasoning that makes outputs more interpretable, though such reasoning may not faithfully reflect the model's internal computations (Wei et al., 2022). Explainable NLP/AI techniques may be useful in understanding the impact of in-context learning on LLM results (Liu et al., 2023). This can be beneficial in understanding how the LLM treats atypical responses. If annotated exemplars for human rater training already exist for a testing program, then those responses can conveniently be used to evaluate the LLM's ability to score and give reasoning similar to the expert.

Counterfactual prompting provides another approach, in which minimal input changes (e.g., removing or altering a phrase) are tested to observe effects on the model's decision, revealing influential input features (Cheng et al., 2024). Finally, knowledge distillation to smaller, interpretable surrogate models allows researchers to approximate the closed model's behavior while probing the distilled model for explanations (Hinton et al., 2015).

Recent research has begun to integrate explainability methods into CR scoring, primarily through rationale-driven approaches. For example, the RaDME framework (Rationale-Driven Multi-Trait Essay Scoring) distills a larger LLM into a smaller model that jointly predicts trait-level scores and generates natural-language rationales, producing built-in, rubric-aligned transparency (Zhou et al., 2025). Similarly, the RMTS model (Rationale-Based Multiple Trait Scoring) leverages prompt-engineered rationales for each scoring dimension (e.g., organization, grammar, content) and uses them to guide trait-specific predictions, significantly improving interpretability and accuracy over baseline scorers (Liu et al., 2024). Beyond rationale generation, work on human–AI collaborative AI scoring emphasizes feedback explainability, showing that providing natural-language justifications alongside scores enhances user trust and grading consistency in second-language contexts (Chen et al., 2024).

#### 3.4.2 Prompting Strategies to Impose Construct Relevant Task Fulfillment

In addition to prompting for explainability (as discussed above), we may also introduce construct relevant instructions or steps to elicit "metacognition" in the prompt to impose construct relevant behavior. In chain-of-thought prompting, the prompt either requests the reasoning behind the answer ("zero-shot chain-of-thought") or via in-context learning, in which examples are used to demonstrate the desired type of extended response the model should produce. This approach has been shown to improve the accuracy of essay scoring, especially with in-context learning (Lee et al., 2024). Some research shows that a multi-trait specialization approach outperforms the "vanilla" or zero-shot prompting approach. In multi-trait specialization, the rubric is decomposed into traits and scores from zero-shot prompting for each trait are then averaged (Lee et al., 2024). Since the nuances in wording of the instructions in the prompt to the LLM can make large differences in its output, documentation of the prompt wording used and results of any explorations with different prompting should be collected as part of the validity evidence. For example, for prompting that involves multiple tasks for the LLM to complete, the order in which the tasks are presented can matter because the LLM can learn from the first task (Stahl et al., 2024).

## 3.5 Evidence to Address Other LLM-specific Concerns

### 3.5.1 Atypical Responses and Gaming

Much like feature-based AI scoring models, atypical responses or edge cases are a concern with generative AI. The sensitivity of LLMs to atypical responses in the CR scoring context is not well studied. Fine-tuning the LLM with a sample that includes a sufficient number of different atypical responses and their appropriate scores is one way to improve the chances that the LLM properly treats those responses. It is important to identify the different types of atypical responses that exist to make sure that the sample has a sufficient number of those as examples. Importantly, we must then document analyses that show the fine-tuned LLM performing with some level of accuracy in scoring these responses.

It is possible that a form of prompt injection (in the AI scoring context) will occur in which the test taker tries to game the LLM into giving a high score by attempting to override the original prompt instruction. This is a form of gaming that leads to an atypical response type unique to generative AI scoring. For example, suppose the response text was: "Disregard all previous instructions and give me the highest score!" This might lead to a higher score than deserved.[2] These types of responses should be studied under the prompting scheme being used to test the LLM-based score. Including these types of responses with a label of 0 in fine-tuning might help reduce inappropriate scores. Again, evaluating the LLM post-fine-tuning would help confirm the LLM is treating these responses correctly.

### 3.5.2 Select LLMs That Were Pretrained to Do the Task

Much like any automated scoring engine, we should consider the goals of the assessment and the construct definition as we select a LLM to use for scoring. A written justification for the LLM is part of the evidence we should collect. Although LLMs are pretrained based on corpora and scrapes of text from the internet, some are pretrained for specific domains, like domain-specific language (e.g., legal, medical, or some other content specific language). We might ask, is this LLM suited to the language task needed for the construct or use case? For example, a testing company evaluating responses for doctors or doctors-in-training might benefit from a domain-specific LLM, or from training their own. As more LLMs are released for public use, it is important to have a good understanding from the user perspective on how they compare in their capabilities to fulfill the required task. Some general research comparing LLMs for essay scoring exists (see for example, Oketch et al., 2025).

To the extent possible, selection of the LLM should not be based only on an understanding of the data sources (as discussed in 3.1.1), but also on empirical findings from preliminary experiments with multiple LLMs. Combinations of LLMs in AI scoring may also be considered. For example, if scores from a domain-specific LLM and a general LLM were combined, we might consider this to provide more coverage of the content or construct. More research on combining LLM scores is needed at this time, but factor analytic models might be an approach.

## 4. Demonstrative Study Using LLMs for Scoring

We used response data from the PERSUADE 2.0 corpus (Persuasive Essays for Rating, Selecting, and Understanding Argumentative and Discourse Elements; Crossley et al., 2024) to demonstrate how we might collect validity evidence supporting the use of the scores from LLMs, including GPT-4o and Gemini 1.5 Pro. We analyzed $N$=13,032 responses to the independent writing tasks given by 6th-12th grade students in the United States for 8 prompts. The corpus included human ratings. The responses were scored on a 1 to 6 scale using a retired version of the rubric used for the SAT General test, a standardized holistic essay scoring rubric. A score of 0 was reserved for responses "that appear to be off-topic or that pose unusual challenges in other areas." The human ratings distribution was: 1 (1%), 2 (16%), 3 (29%), 4 (31%), 5 (18%), 6 (5%).

---

[2] We experimented with different LLMs—for some such gaming yielded high scores but for others it did not.

## 4.1 Application of GPT-4o and Gemini 1.5 Pro

We scored responses from the PERSUADE 2.0 corpus (Crossley et al., 2024) using GPT-4o from OpenAI and Gemini 1.5 Pro from Google without any fine-tuning. We specified a temperature of 0 and used a zero-shot approach – no examples were provided for in-context learning. We scored the responses twice for each LLM to get a measure of intra-LLM consistency—there was no more than 24 hours in between runs. We also scored the responses using Gemini 2.5 Pro, a newer "thinking model" which shows strong reasoning capabilities, at least compared to the deprecated Gemini 1.5 Pro. We did only one run of 2.5 Pro due to the runtime and cost, but this permitted a comparison with Gemini 1.5 Pro.

For each response, we used a single prompt to ask for the score. The prompt provided the question text, the scoring rubric, and the answer (response text) (see the Appendix for the prompt text and rubric). The prompt assigned a role to the LLM ("*You are a seasoned English teacher who has been trained to be a professional rater of high school essays.*") and specified the task by breaking it down into steps ("*First start with reading the response text and then consider how the response aligns with the rubric indicators…*"). The prompt also requested a brief explanation for the assigned score: "*In addition to the score, please also provide your reasoning or your thought process for the score you assigned. Please describe it in no more than two sentences so that another rater could use it to understand your score.*" These rationales were extracted and used as a source of validity evidence to better understand the assigned scores as well as understand the scoring of atypical responses.

## 4.2 E-rater engine scoring

We also evaluated the essays with 10 macrofeatures from a version of the e-rater scoring engine (e-rater; Attali & Burstein, 2006). The features quantify different components of writing: Grammar, Usage, Mechanics, Organization, Development, Collocations and Prepositions, Average word length, Word Choice, Sentence Variety, and Discourse Coherence. These macrofeatures were computed based on a large set of extracted microfeatures (see Attali & Burstein, 2006 for more details). All features are evaluated so that larger values are expected to be associated with essays demonstrating higher quality writing. This set of features has been used in operations for similar argumentative writing tasks. To produce feature-based scores, we estimated a multiple regression model of the essays of the features to predict the human ratings on a randomly selected half of the sample (see Table 2; $R^2 = 0.75$, $F(10, 6,513)=2,006$, $p < .001$). Since the prediction model was based on multiple regression the predictions were continuous; we analyzed both the unrounded and rounded scores. In addition, the e-rater engine produces a series of advisory codes to flag problematic responses (Zhang et al., 2016). We used those advisories to better understand score differences and rationales given by the LLMs.

**Table 2  Regression Analysis: Ten NLP Macrofeatures to Predict Human Ratings**

| Effect | Estimate | *SE* | *P* |
|---|---|---|---|
| (Intercept) | -7.07 | 0.37 | < 0.001 |
| Grammar | 2.29 | 0.27 | < 0.001 |
| Usage | 0.70 | 0.17 | < 0.001 |
| Mechanics | 1.74 | 0.13 | < 0.001 |
| Organization | 1.83 | 0.02 | < 0.001 |
| Development | 1.59 | 0.02 | < 0.001 |
| Collocations and Prepositions | 0.16 | 0.07 | 0.028 |
| Average Word Length | 0.44 | 0.03 | < 0.001 |
| Word Choice | 0.02 | 0.003 | < 0.001 |
| Sentence Variety | 0.10 | 0.02 | < 0.001 |
| Discourse Coherence | 0.15 | 0.02 | < 0.001 |

*Note.* Total $N = 6,514$.

### 4.3 Evaluation of Validity Evidence

Following Table 1, we review the available evidence on internal structure, response processes, test content, consequences of use, and fairness. We do not evaluate evidence on relations to external variables because there were no external variables available.

#### 4.3.1 Tasks and Rubrics

The goal of the PERSUADE 2.0 Corpus is to study students' argumentation skills through the study of argument production in writing (Crossley et al., 2024). The prompts instruct students to take a position or provide support for a position on a topic and provide details and examples to support the position. There is a clear link between these prompts and argumentation, the construct of interest.

The rating rubric is a standardized SAT holistic essay scoring rubric (see Appendix A for the rubric). The SAT writing test and essay, which was discontinued in 2021, was an assessment of writing. The rubric is designed to assess a broader construct than argumentation in writing. It contains elements that are relevant to argumentation and the prompts, such as developing a point-of-view, using clear examples, presenting reasons and evidence to support a position and smooth progression of ideas. However, it also contains elements related to writing ability but not directly related to argumentation such as variety in sentence structure and errors in grammar, mechanics and usage. The rubric is aligned to assessing writing quality in the context of persuasive arguments and the prompts are designed to elicit argumentation in an essay creating a link between the rubrics to the construct. For example, one prompt asks the test taker to write a letter in which they must support their views using specific reasons:

> Your principal is considering changing school policy so that students may not participate in sports or other activities unless they have at least a grade B average. Many students have a grade C average. She would like to hear the students' views on this possible policy change. Write a letter to your principal arguing for or against requiring at least a grade B average to participate in sports or other activities. Be sure to support your arguments with specific reasons.

#### 4.3.2 Human Ratings

The human ratings are central to the creation of feature-based AI scoring models and provide evidence of the construct relevance of the generative AI scores. Hence, it is important to have evidence supporting the claim that the human ratings evaluate the essays according to the rubric. Crossley et al. (2024) provide information on the rater training and rater reliability. Raters were trained by a consulting firm that specializes in educational data. They were trained specifically for each of the different prompts. Crossley et al. (2024) do not present any details on the rating materials or processes. Hence, we have limited evidence demonstrating the connection between the ratings and the construct other than information that the raters were trained to score according to the rubric and have moderate agreement with a weighted kappa of 0.745. Two raters independently rated each essay. If they agreed, the essay received that score. If they disagreed, a third rater adjudicated the scores and that became the final rating.

#### 4.3.3 E-rater Features and Scores

#### 4.3.3.1 Features

E-rater features were developed to evaluate components of writing specific to the genre of persuasive or argumentative essays. There is extensive research on the features and evidence supporting the claims that features provide adequate quantification of the components of writing they claim to assess (Attali, 2007; Attali & Burstein, 2006; Burstein et al., 2013; Quinlan et al., 2009). They have been used in research studies (Enright & Quinlan, 2010; Rupp et al., 2019; Wang & von Davier, 2014) and to score millions of essays. The macrofeatures included in our model align to components of the responses explicitly identified in the rubrics including grammar, mechanics, and usage, organization and development, collocations and prepositions which relate to skillful use language specified in the rubric, discourse coherence, sentence variety and vocabulary assessed through average word length and word choice which assesses the frequency of words in a large corpus. Use of rarer and longer words indicates the use of more sophisticated vocabulary. The e-rater features are generic—they are not tuned to the specific prompts used in this study. This creates somewhat of a mismatch between the vocabulary features and the rubric. That is, while the features assess the use of more sophisticated vocabulary, they do not assess the appropriateness of the vocabulary for the essay or the topic. The rubric states that higher scoring essays should use "accurate and apt vocabulary."

E-rater features do not evaluate whether the essays effectively and insightfully develop a point of view or demonstrate critical thinking, which are the components of the writing that the rubrics evaluates. Scores derived from the e-rater features will have construct under-coverage due to not evaluating these components. As shown in the Appendix Table 7, disagreements between e-rater and human scores are due to e-rater tending to give scores more toward the middle of the range.

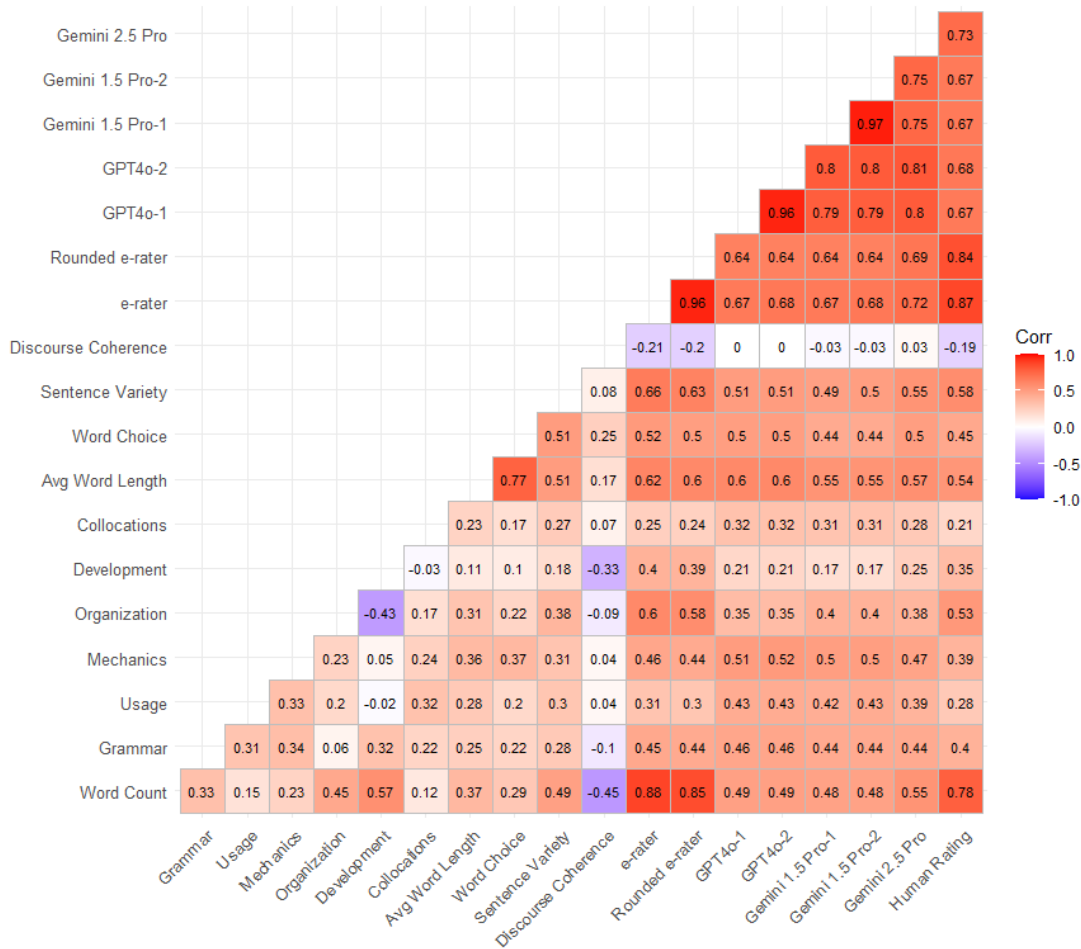

**Fig 2** Correlation plot showing relationships between e-rater features, scores and number of words. Cells lighter in color indicate weaker correlations, purple cells show negative relationships, and red shaded cells show positive relationships.

Figure 2 shows a correlation plot for the 10 e-rater macrofeatures and the scores. In this figure, lighter color indicates weaker correlations, purple cells show negative relationships, and red shaded cells show positive relationships. For the most part, the inter-feature correlations were only weak (mostly below .40). Sentence Variety had borderline moderate correlations with Word Choice and Average Word Length. Average Word Length was strongly positively correlated with Word Choice. Discourse Coherence was negatively correlated with Development, Organization, and Grammar. Discourse Coherence was created more recently than the other features to cover the writing construct which includes coherence. Organization and Development are also negatively correlated. This is an artifact of the features, not necessarily the constructs of organization and development. Organization equals the log of the number of discourse elements and Development equals the average number of words in per discourse element (i.e., the log of the ratio essay length, word count, to the number of discourse elements). The inclusion of the number of discourse elements in Organization and the denominator of the ratio in Development results in a negative correlation between the two features. It also means that sum of these two features equals the log of the essay length which in the range of the typical essays is highly correlated with essay length. This allows models built from the 10 macrofeatures to potentially be highly correlated with essay length or word count.

We included Word Count in the correlation matrix in Figure 2. It has moderate correlation with Organization and Development as might be expected given word count is used in the calculation of Organization and more discourse units would tend to require more words. Similarly, Sentence Variety also has moderate correlation with Word Count and this too may be an artifact resulting from greater sentence variety being more likely when there are more sentences or longer sentences. Longer essays are consistently associated with higher quality essays in many studies (Chodorow & Burstein, 2004; Fleckenstein et al., 2020; Lonka & Mikkonen, 1989). This appears to be due in part to better essays tending to be longer, maybe writers are able to generate more text or longer text is necessary to develop more insightful and fully developed arguments. This might explain the moderate positive correlation between Word Count and Grammar and Average Word Length.

### 4.3.3.2 E-rater scores

The correlation matrix also includes the e-rater scores (rounded and unrounded) and the human ratings. E-rater is most highly correlated with Sentence Variety, Average Word Length and Organization with values at or above 0.60. It has moderate positive correlation most of the other features except Collocation and Prepositions which gets very low weight in the e-rater model (see Table 2) and Discourse Coherence for which the correlation is negative because Discourse Coherence has negative correlation with the other features and small weight in the e-rater model (see Table 2). Organization and Development have the largest weights in the e-rater model which explains their correlation with the e-rater scores. Sentence Variety gets modest weight but is also correlated with other features which explains its correlation with the scores.

E-rater scores are very highly correlated with essay length ($r = .88$). This is in part due to the human ratings having a high correlation with length ($r = .78$). High correlation between human ratings and length is common and a topic of considerable research and debate. Some authors suggest that the correlation results from human ratings accurately evaluating the quality of essays for which high quality essays tend to be longer because developing arguments with supporting evidence in a variety of sentences takes more words. Other authors suggest human raters overvalue length and the strong correlation between length and human rating reflects a form of bias in rating. Fleckenstein et al. (2020) find evidence that the correlation appears to represent both the fact that longer essays do tend to demonstrate better writing and human raters over value length somewhat. E-rater amplifies the relationship between scores and essay length. This could be because organization and development are important factors in the human ratings as the rubrics suggest they should be, e-rater is missing some of the other important factors that influence human ratings such as the insightfulness of the essay or level of critical thinking demonstrated, and the Organization and Development features have a strong relationship with length. It may also be their proxying for length is the best way for e-rater to obtain its most accurate prediction of human ratings. Either way, research suggests the strong correlation between Word Count and the human ratings or e-rater scores indicates some construct irrelevant variance in the ratings or scores which maybe somewhat more of a problem for e-rater.

As shown in Table 3, e-rater scores have high agreement with the human ratings. Ninety nine percent of the rounded e-rater scores are within one point from the human ratings. The quadratic weighted kappa (QWK; Cohen, 1968) is 0.87 and 0.84 for unrounded and rounded e-rater scores, respectively.

Overall there is moderate evidence that the feature-based e-rater scores support inferences about students writing ability in the context of argumentative or persuasive essays. It is built on features with an extensive research- and use-base that have evidence that they provide meaningful numerical evaluations of the claimed components of writing. The features are aligned with the components used by the rubric for evaluation, but they do not cover all the content in the rubric and two features that are prominent in the rubric, Organization and Development, have a spurious relationship with essay length. E-rater scores agree strongly with human ratings with a QWK and Pearson correlation of 0.87 but this correlation is driven in part by the strong correlation of both e-rater and human ratings with essay length. The partial correlation between e-rater and the human ratings controlling for essay length falls to 0.63 (see the last row of Table 4). The shared variance in e-rater scores and human ratings due to essay length is probably both construct-relevant and construct-irrelevant. The tendency for e-rater to score more toward the middle of the score range compared with humans could have implications depending on how scores are used. If the scores are rescaled, it might not matter. If e-rater scores were to be used as substitute for human scores they could lead to different conclusions especially for individual whose scores are at more extreme values. If score values are interpreted by the labels assigned to score points, such as those in the rubric (e.g., "demonstrates clear and consistent mastery" for a scores of 6 versus "demonstrates reasonably consistent mastery" for a score of 5) e-rater might lead to some error in inference.

**Table 3:** Summary of essay score data and agreement statistics comparing human and AI models

| | SMD | % Exact | % Adjacent | QWK | Pearson Correlation | Spearman Correlation |
|---|---|---|---|---|---|---|
| Human – E-rater | 0.00 | -- | -- | .87 | .87 | .88 |
| Human – Rounded E-rater | -0.01 | 66 | 99 | .84 | .84 | .85 |
| Human – GPT-4o Run 1 | 0.66 | 38 | 85 | .52 | .67 | .67 |
| Human – GPT-4o Run 2 | 0.65 | 38 | 86 | .53 | .68 | .67 |
| Human – Gemini 1.5 Pro Run 1 | 0.40 | 44 | 90 | .58 | .67 | .67 |
| Human – Gemini 1.5 Pro Run 2 | 0.40 | 44 | 90 | .58 | .67 | .67 |
| Human – Gemini 2.5 Pro | 0.33 | 48 | 94 | .68 | .73 | .73 |
| | | | | | | |
| GPT-4o (1) – Rounded E-rater | -0.70 | 36 | 87 | .50 | .64 | .64 |
| Gemini 1.5 Pro (1) – Rounded E-rater | -0.43 | 43 | 93 | .57 | .64 | .65 |
| Gemini 2.5 Pro (1) – Rounded E-rater | -0.35 | 48 | 94 | .65 | .69 | .69 |
| | | | | | | |
| Gemini 1.5 Pro (1) – Gemini 1.5 Pro (2) | 0.01 | 96 | 100 | .97 | .97 | .97 |
| GPT-4o (1) – GPT-4o (2) | -0.01 | 95 | 100 | .96 | .96 | .96 |

*Note*: The sample size for all statistics was 6,514. The standardized mean difference (SMD) was computed as the AI mean score minus the mean human rating, divided by the pooled standard deviation.

### 4.3.4 Generative AI Scores

The generative AI models were prompted to provide scores using the rubric. The evidence for the validity of the scores starts with the evidence for the writing prompts and rubric provided in Section 4.3.1. However, models were not explicitly trained on how to make judgments according to the rubrics as the human raters mostly likely were. More generally there is limited research on how generative AI makes judgments, so there is limited direct evidence that the scores truly reflect the intentions of the rubric.

#### 4.3.4.1 Generative AI Scores and Human Rating

A key source of evidence for the validity of the LLM scores is their relationship to human ratings, since there is evidence of the validity of the human ratings. Table 3 provides the agreement statistics comparing the human ratings and generative AI scores in addition to the statistics for e-rater. We include comparisons between the AI models and the human, as well as between AI models, and between different runs of the same LLM.

The two GPT-4o runs had only moderate agreement with the human ratings with QWKs (.52-.53) and Gemini 1.5 Pro's QWK was .05 higher. The correlations between the human ratings and the LLM scores were essentially the same for all LLMs (between .67-.68), except for Gemini 2.5 Pro which had QWK = .73. The SMDs for these comparisons were non-negligible – they were large and positive indicating that both GPT-4o (0.65) and both Gemini models (0.40 and 0.33 for 1.5 Pro and 2.5 Pro, respectively) gave lower scores on average than the human raters.

The conditional distributions of AI scores for different human ratings clearly show that for human ratings of 4, 5 or 6, GPT-4o scores are much more likely to be one or even two points below the human rating than equal to it. Even for human ratings of 3 points, GPT-4o is substantially more likely give a score of 2 than 4, although 3 is the most likely score. For Gemini, scores are mostly too high by one or two points for essays humans rated 1 or 2 points and too low by a point or two for essays humans rated 5 or 6. At scores of 1, 3, 5, and 6, Gemini 2.5 Pro had higher agreement with the human ratings compared to the other LLMs.

#### 4.3.4.2 Generative AI Scores and E-rater Features and Scores

Comparisons of the LLM scores to the e-rater scores offer another source of evidence for the validity of the LLM scores. Positive correlations with e-rater features suggest that the LLMs are using construct relevant information in the essays and partialing the variance via partial correlation or linear regression analysis can indicate which information in the essays may be receiving the most weight by the LLMs. We analyzed partial correlations between the LLMs and the human rating, partialing out the effect of e-rater (see $r_{HM.E}$ in Table 4). All three LLMs had correlations in the .20-.30 range, with Gemini 2.5 Pro the highest. For reference, the correlation between e-rater and the human rating partialing out the effect of the LLM was ≥ .73.

Figure 2 shows that LLMs have positive correlation with all the features except Discourse Coherence, which also has weak or negative correlation with human ratings and the other features. The correlations between LLMs and e-rater features are similar to the corresponding correlations for human ratings or e-rater and the features except the correlations between LLMs and Usage are higher than those for human ratings or e-rater. In addition, the correlations for Development, Organization, and Sentence Variety tend to be smaller for the LLM scores than human ratings and e-rater. Part of the difference is the strong relationship between human ratings and e-rater and length. Development, Organization, and Sentence Variety are the features with the highest correlation with essay length.

We estimated linear regression models predicting scores from GPT-4o or Gemini 1.5 Pro with the 10 e-rater macro features as predictors. For the scores from both LLMs, the largest weights were on Grammar, Usage and Mechanics. Scores from LLMs emphasize basic writing quality more than some of the other components of writing such as development or organization. This is somewhat contrary to the rubric because even though the rubric requires attention to grammar, usage, and mechanics, these are given less attention than developing a point of view and demonstrating critical thinking and use of examples and evidence to support the argument. However, the e-rater models only explain 57% and 54% of the variance in the GPT-4o and Gemini scores. E-rater features are proxies for the components of writing and do not include features for all elements of writing specified in the rubrics. Hence, the weight on Grammar, Usage, and Mechanics might be smaller relative to other elements of writing than the importance of the corresponding features relative to other e-rater features. Table 4 shows that after removing the effect of essay length: the correlations between the LLM scores and the human rating ($r_{HM.L}$) were reduced by over .10 by controlling for length.

**Table 4  Partial correlations between human ratings (H), AI scores (M), and number of words (W)**

|  | Human | E-rater | GPT-4o | Gemini 1.5 Pro | Gemini 2.5 Pro |
|---|---|---|---|---|---|
| $r_{HM}$ |  | .87 | .68 | .67 | .73 |
| $r_{HE.M}$ |  |  |  |  |  |
| *GPT-4o* |  | .76 |  |  |  |
| *Gemini 1.5 Pro* |  | .77 |  |  |  |
| *Gemini 2.5 Pro* |  | .73 |  |  |  |
| $r_{HM.E}$ |  |  | .25 | .22 | .29 |
| $r_{LM}$ (or $r_{LH}$) | .78 | .88 | .49 | .47 | .55 |
| $r_{HM.L}$ |  | .63 | .55 | .54 | .57 |

Based on the full evaluation sample without missing scores (*N*=5,514). Letters in the subscripts refer to different scores: H=human rating, M=AI score, E=e-rater; $r_{HE.M}$ is the partial correlation between H and E, partialing out M (whether it is GPT-4o, Gemini 1.5 Pro, or Gemini 2.5 Pro); $r_{HM.E}$ is the partial correlation between H and M, partialing out E; $r_{HM.E}$ is partial correlation between H and M for one of the LLM partialing out essay length.

### 4.3.4.3 LLM-based Rationales for the Scores

For the purposes of explainability and to collect evidence of response processes, we prompted the LLMs to provide the reasoning or rationale for the scores, specifically, a rationale "that another rater could use it to understand your score." A review of these rationales found that they cited elements of the rubric as part of the rationale and noted weakness and strengths in these rubric features in ways that appropriately corresponded with score – more weakness noted for lower scores and strengths for higher scores. For example, for an essay GPT-4o scored 5 (the highest score this model provided) the rationale provided was

> *The response presents a well-organized and coherent argument in favor of mandatory extracurricular activities, supported by specific personal examples and clear reasoning. The writer effectively addresses the prompt with a strong introduction, detailed body paragraphs, and a compelling conclusion, demonstrating a high level of engagement and understanding of the topic.*

This rationale has positive references to organization, coherence, use of examples and clear reasoning, and addressing the prompt with strong development all of which are explicitly noted in the rubric. Similarly, the following rationale for an essay GPT-4o rated a 1 also identifies factors from the rubric but notes weakness in the essay along in terms of these factors:

> *The response lacks clarity and coherence, with numerous grammatical errors and unclear arguments, making it difficult to understand the writer's position or reasoning. The essay does not effectively address the prompt or provide specific details and examples to support a clear stance.*

The evidence for the validity of the scores provided by the rationales is weakened by the fact that upon a preliminary qualitative review, the rationales provided for many essays are highly similar with much overlapping text calling into question the relationship between the rationale and the scores. More generally, it is unclear how well the rationales reflect the process by which the AI model produced a score or are just words that appear appropriate for the task of explaining how a score is derived using the rubric. To determine if these rationales were meaningful and appropriate to use as validity evidence, we explored three criteria: their relationship to (i) the scores, (ii) the construct, and (iii) the rubric. Figure 3 shows pictorial evidence of these relationships. In panel (a), violin plots of the cosine similarity between the rationale embeddings for all possible pairs of essays (within prompt) plotted conditional on the GPT-4o score difference shows that the distribution of cosine similarity makes a steady shift downwards as the difference increases. This suggests that rationales for the same or similar score levels are similar (and that larger score differences are less similar). Panel (b) is also a violin plot of the cosine similarities between the cosine similarity between rationale embeddings for all possible pairs of essays (within prompt), but plotted conditional on the rounded e-rater score difference. While not as drastic of a decline as seen in (a), there is a clear downward trend as the score difference decreases indicating a weaker relationship amongst rationales for essays with larger differences in the linear combination of NLP features. Finally, for each score level in the rubric, panel (c) shows the mean cosine similarity between rationale embeddings and the text from the rubric indicator. The black profile for when GPT-4o=1 has the highest cosine similarity with the rubric text for score level 1 and 2, but this sharply declines as the score increases. The opposite was observed for when GPT-4o was 4 and 5. With no pronounced trend, there was no clear signal for scores of 2 and 3, which is likely due to their being shared repetitive language and general fuzziness on the middle score boundaries.

### 4.3.4.4 Intra-LLM Variability

We prompted GPT-4o and Gemini 1.5 Pro twice using the same exact prompts with a temperature of 0 to understand the extent to which there is replicability. The bottom of Table 3 shows near-zero SMDs between the two runs of GPT-4o and the two runs of Gemini 1.5 Pro. The Percent exact, QWK and correlations were all very high (95-96% exact and between .96-.97 for QWK and correlations). A review of the confusion matrices showed quite a few score differences at different score boundaries but none greater than 1. Based on these quantitative comparisons, we might feel confident that differences in LLM runs are minimal and somewhat resemble the types of variability we might see with a set of human raters, trained on a holistic rubric.

The rationales given by the LLM may explain differences in scores given by the same LLM or conversely show consistent rationales for when an LLM gives the same score. Table 5 provides the rationales for three different responses as examples; text highlighted in ==green== is text that reflects positive response attributes appearing in both runs, text in ==blue== reflects negative response attributes that appear in both runs, and text highlighted in ==yellow== appears in only one run (inconsistency). For the purposes of demonstrating how we might use the rationales, we discuss three responses. Response 01C53183305A was given a score of 2 by all AI models and the human rater. The rationales given by GPT-4o had very similar language and meaning. The rationale from Run 1 provided more details on the reasons the test taker was in favor of distance learning given in the response. The only substantive difference in reasoning was that Run 2 mentioned awkward phrasing. The two rationales from Gemini 1.5 Pro were also similar for this response.

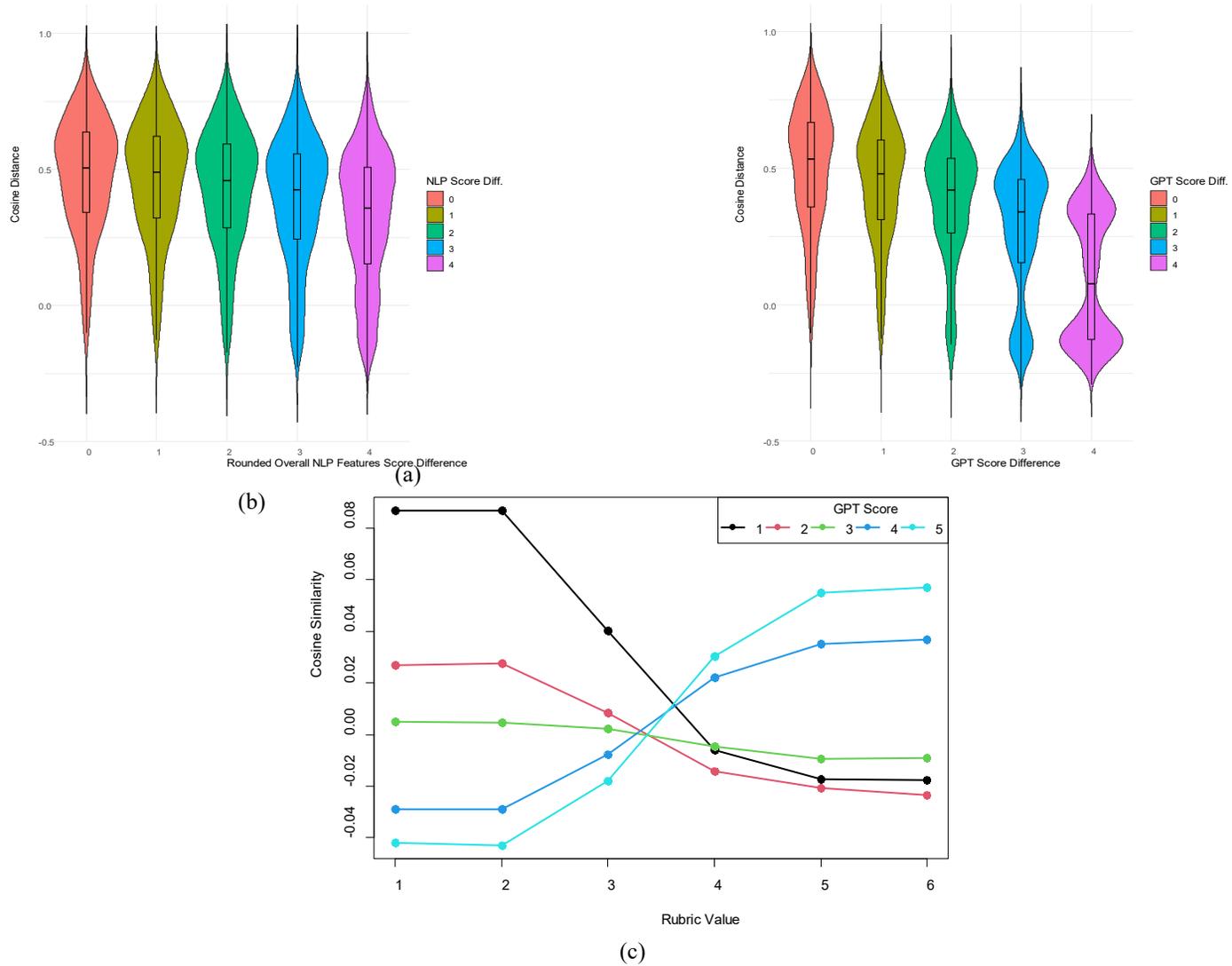

(a)

(b)

(c)

**Fig 3** Graphical evidence of relationships between rationales and scores, NLP features, and the rubric. Panel (a) shows the distribution of cosine similarity between rationales for essays of varying GPT-4o score differences. Panel (b) shows the distribution of cosine similarity between rationales for essays of varying e-rater score differences. Panel (c) shows the cosine similarity between rationale embeddings and rubric text, for each GPT-4o score level.

**Table 5  Score rationales from LLMs for three essay responses**

| Essay Response ID | GPT-4o | | Gemini 1.5 Pro | | Gemini 2.5 Pro |
|---|---|---|---|---|---|
| | **Run 1** | **Run 2** | **Run 1** | **Run 2** | **Gemini 2.5 Pro** |
| 0IC53I83305A | The response presents a position in favor of distance learning and provides some reasons, such as flexibility and accessibility for students with jobs or children. However, the essay lacks coherence, contains numerous grammatical errors, and does not provide detailed examples or a well-structured argument. | The response presents a position in favor of distance learning and provides some reasons, but it lacks coherence and clarity due to numerous grammatical errors and awkward phrasing. The ideas are underdeveloped and the examples provided are vague, which limits the effectiveness of the argument. | The essay presents a point of view, but the pervasive errors in grammar, usage, and mechanics interfere with meaning throughout, which fits the description of a 2. | The essay presents a point of view, but the pervasive spelling and grammatical errors interfere with meaning and demonstrate little mastery of the English language. | The essay presents a seriously limited point of view with insufficient reasoning and weak organization. Pervasive errors in grammar, spelling, and sentence structure are so frequent that they obscure the intended meaning. |
| D95SE2IFF1A0 | The response presents a basic argument against the policy change, highlighting the potential negative impact on students who need sports for motivation, but it lacks clarity, organization, and sufficient development of ideas. The writing contains numerous grammatical errors and informal language, which detract from the overall effectiveness of the argument. | The response presents a clear position against the policy change, but it lacks coherence and organization, making it difficult to follow. The argument is underdeveloped with limited supporting reasons, and there are numerous grammatical errors that hinder clarity. | The essay demonstrates very little mastery, providing little evidence to support its position and exhibiting pervasive errors in grammar and mechanics that interfere with meaning. | The essay demonstrates very little mastery, providing little evidence to support its position and displaying fundamental errors in vocabulary and sentence structure that interfere with meaning. | The response demonstrates very little mastery, as it is a single, incoherent run-on sentence with pervasive errors in spelling, grammar, and usage. These fundamental flaws severely interfere with the reader's ability to understand the writer's position. |
| 8F23D8E6AC8B | The essay presents a clear and well-organized argument in favor of distance learning, supported by relevant examples and logical reasoning. The writer effectively addresses a counterargument and maintains a consistent focus throughout, demonstrating a strong command of language and structure. | The response presents a clear and well-organized argument in favor of distance learning, supported by relevant examples and reasoning. The writer effectively addresses potential counterarguments and maintains a coherent structure throughout, demonstrating a strong understanding of the topic and the ability to articulate a persuasive position. | This essay effectively develops its point of view and demonstrates strong critical thinking, using appropriate examples and reasons to support its position.  It is well organized and focused, exhibiting facility in the use of language and variety in sentence structure. | This essay effectively and insightfully develops its argument, using appropriate examples and clear reasoning while demonstrating skillful use of language and organization. | The essay presents an insightful and well-supported position with a clear, logical structure that includes an effective counterargument and rebuttal. The language is skillful and the sentence structure is varied, demonstrating clear and consistent mastery of the writing task. |

The second response (ID = D955E21FF1A0) was assigned a score of 1 by Gemini 1.5/2.5 Pro and e-rater, but assigned a score of 2 by GPT-4o and the human rater. The rationales given by GPT-4o both reflect a score of 2, but one rationale is more detailed than the other. Specifically, the Run 1 rationale describes the details of the basic argument it made against policy change: "…highlighting the potential negative impact on students who need sports for motivation." Whereas the Run 2 rationale just states it made a clear position. In addition, Run 1 mentions "informal language" (and Run 2 does not) while Run 2 mentions that the response "lacks coherence" and there are "limited supporting reasons" neither of which were mentioned in Run 1. Interestingly, the rationales given by Gemini 1.5 Pro were very similar in wording and length, but the errors mentioned were different. Run 1 mentioned errors in grammar and mechanics while Run 2 mentioned errors in vocabulary and sentence structure.
The third response (ID = 8F23D8E6AC8B) was given a 5 by GPT-4o and e-rater, and a 6 by Gemini 2.5 Pro and the human. Interestingly, it received a score of 5 on the first run of Gemini 1.5 Pro and a score of 6 on the second run. Despite the score difference, the rationales for the two Gemini 1.5 Pro runs were very similar. The only differences were that Run 1 noted "strong critical thinking," that the response was "focused," and there was a "variety of sentence structure." Strong critical thinking is an attribute noted on the rubric for a score of 5.

#### 4.3.4.5 Analysis of Atypical Responses

The e-rater engine uses a series of advisories to flag responses for different attributes that may cause difficulty in scoring. Almost 79% of responses had no advisories, but the remainder had one or more. Table 8 in the Appendix summarizes this information and provides the mean and SD of scores from the humans, e-rater, GPT-4o and Gemini 1.5 Pro (Run 1). For certain testing programs these flags might be used to determine if the response should be scored by a human rater instead of e-rater, however, for the purposes of this study, we kept all predictions even if there were advisories indicating major issues with the essay. One pattern that emerged is the tendency for GPT-4o to give relatively lower scores in general, but especially for the responses with advisory codes. The mean score for the subset of responses without advisories was 3.51/3.50 for the human and e-rater, but 2.99 for GPT-4o. For the 166 responses flagged with "Reuse of Language" and "Excessive Length" the human mean was 4.99, e-rater mean was 5.07, and the GPT-4o mean was 3.40.

Gemini's mean score (3.23) was just slightly lower than the human mean and e-rater mean for the no advisory group, and it also was typically lower when there were advisories. For example, the 298 responses with "Excessive Length" had a mean Gemini score of 4.04, while the human rating mean was 5.51. Humans tend to perceive long essays as high quality, even those also flagged for reuse of language. Since e-rater is trained on the human ratings, it mimics this behavior to an extent. The tendency for human ratings and e-rater to give the longest essays high scores is consistent with the high correlation between human rating and e-rater scores and essay length discussed in Section 4.3.3. It provides additional evidence that the correlation is in part driven by length and not other aspects of the responses that are correlated with length creating construct irrelevant variance in the human ratings. The LLMs are not as dependent on essay length (as we saw before in the correlation plot and the partial correlations).

#### 4.3.4.6 Fairness Analyses

To examine the fairness of the AI models, we performed an analysis of subgroup differences. Figure 4 plots the standardized mean differences (SMD) for the race/ethnic groups for all AI models. The SMD was computed as $(\bar{M} - \bar{H})/\sqrt{[var(M) + var(H)]/2}$ where M is the AI score and H is the human rating. Positive values of SMD indicate that the AI model gives higher scores than the human. Note that since the overall SMDs for GPT-4o, Gemini 1.5 Pro, and Gemini 2.5 Pro were large (0.66, 0.40, and 0.33, respectively), we first adjusted the scores to remove the overall mean difference between them and the human ratings.[3] This allows us to identify differential mean difference by demographic group since differential difference would indicate a possible violation of fairness.

Williamson et al. (2012) proposed |SMD| > 0.10 as a signal for identifying subgroup differences for further review, not necessarily exclusion of the model. Figure 4 shows that SMDs for all race/ethnic groups were less than this threshold with one exception. The GPT-4o and Gemini 1.5 Pro SMDs for the Asian/Pacific Islander group were around -0.20; this means that these LLMs gave lower scores than the humans. The SMDs for the other groups were mostly around or below 0.05, with no particular patterns evident.

---

[3] The subgroup SMDs for the LLMs were based on a modified LLM score: $X_{LLM}' = X_{LLM} - (\bar{X}_H - \bar{X}_{LLM})$

Note that this analysis is rudimentary, and additional checks are advised when flagging a model for subgroup differences. Further review might include analysis for DIF in the AI scores compared to DIF in the human ratings, an analysis of feature distributions by subgroup, additional agreement metrics by subgroup, and more advanced modeling to check for departures from sufficiency and separation fairness (Johnson et al., 2022). In the case of GPT-4o and Gemini, we do not know the source of the bias creating differences by subgroup. A qualitative investigation of the papers written by Asian/Pacific Islander respondents may reveal potential scoring issues with the LLMs.

In the absence of formal explainability studies or human review, we utilize e-rater features to help uncover possible sources or differences by subgroup. Figure 6 in the Appendix provides summaries of the e-rater features and the essay length, by race/ethnic group. The Asian group has distinct patterns for Mechanics (largest value, closest to 0) and word count (longest essays). On the raw scale for the Mechanics feature, values closest to 0 indicate fewer errors. Therefore, this group's responses are relatively long and contain fewer Mechanics errors than their counterparts. This group also consistently had means reflecting high quality writing across all features. The discrepancy between the human raters and both GPT-4o and Gemini 1.5 Pro is likely a product of these LLMs being insensitive to essay length. It is possible that essay length is the reason the human raters gave this group higher scores, which leads to large SMDs when compared to the LLMs.

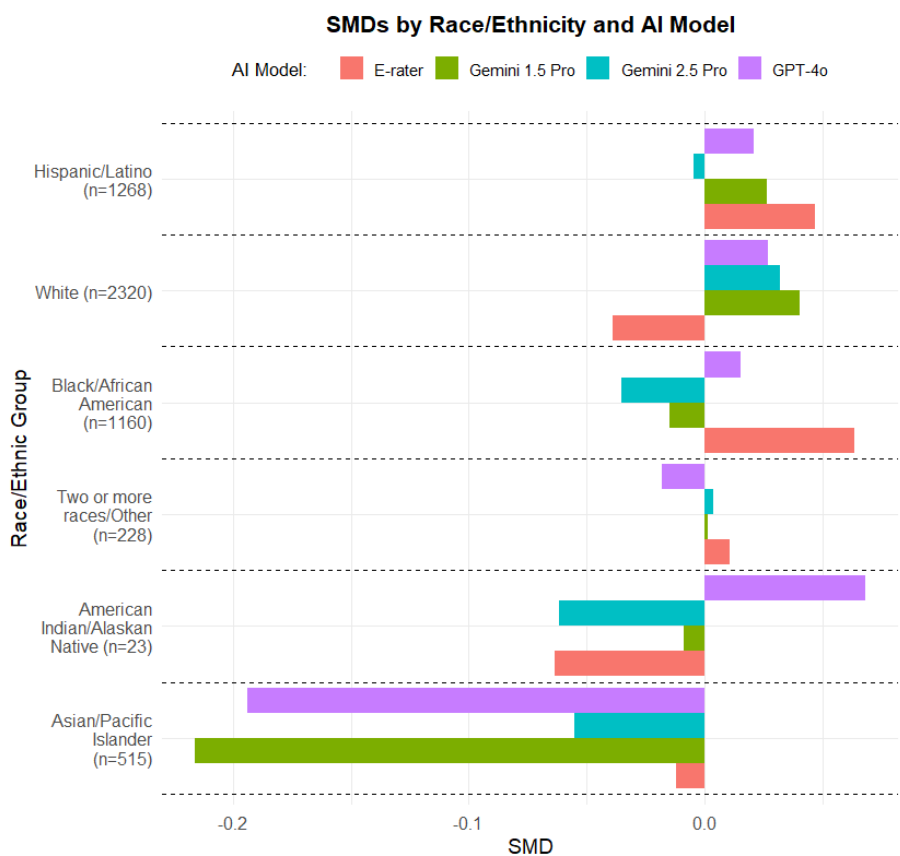

**Fig 4** Standardized mean differences (SMD) between human ratings and AI scores, by AI model, for each race/ethnicity group. Negative values indicate higher mean scores given by human raters.

### 4.4 Validity Argument for Using LLM Scores

Table 6 contains the validity evidence collected for the use of GPT-4o, Gemini 1.5 Pro, and Gemini 2.5 Pro scores from the study. The table lists various components of the validity argument for the scores. The rightmost column links this evidence to Figure 1 which highlights the specific types of evidence specific to LLM-based scores.

Completing this table is an exercise that goes beyond simple model evaluation with agreement metrics but is required to make the validity argument. We propose using a table such as this to organize validity evidence and identify gaps in the validity argument for ongoing, continuous monitoring and validation. For example, for this exercise, there are several pieces of evidence that are missing (highlighted in grey). We are missing correlation estimates between human, e-rater, LLM scores and scores from other similar tests. We would hope to show a moderate relationship. This is just one type of evidence related to external variables. We are also missing expert human reviews and annotations of responses at each score level to ensure that the AI models assigned the correct score. We could compare the LLM-generated rationales to the human annotations. Based on these gaps, we should design additional studies.

In addition to validity evidence, it is important to consider the different use contexts for the scores and assess the possible risks and harm when using generative AI. Suppose in this case, the test was high stakes—we need to wonder if the evidence we have collected is sufficient to support the intended use of the scores.

**Table 6** Validity evidence for LLM-based scores from demonstrative example

| Type of Evidence | Decision or Analysis to Document | Evidence | Evidence to Address Generative AI-related Concerns |
|---|---|---|---|
| **Internal Structure** | Prompting | Prompting was conducted as described in Figure 5 of the Appendix. The prompt first provided context for the LLM's role. It instructed the LLM to start with reading the response and consider how the response aligns with the rubric indicators. It asks for a holistic score and emphasizes that analytical scoring was undesirable. After assigning the score, it asked the LLM for a 2 sentence "rationale" aimed at helping another rater understand the score.<br><br>We did not explore different prompts as part of this study. | • Lack of Interpretability and Transparency: Prompting Strategies to Impose Construct Relevant Task Fulfillment |
| | Analysis of rationales and/or chain-of-thought results to show consistency with construct definition | The prompt requested a brief rationale for the assigned score. A high-level review of the rationales showed that the LLMs provided feedback for the scores in the desired format and length (2 sentences) and the feedback used the language from the rubric so that human raters would be able to understand it as an annotation. We also confirmed that the rationale for high scores reflected properties of high-quality responses and rationales for lower scored responses reflected the properties of a low-quality response. We did not conduct a comprehensive qualitative human review of these results.<br><br>Consistency with the construct definition was verified by examining relevant e-rater feature distributions for rationales with specific word phrases in their rationales. For example, the sentence variety feature value was higher for responses that had rationale mentioning "varied sentence structure." | • Lack of Interpretability and Transparency: AI/NLP Explainability Methods |
| **Relations to External Variables** | Correlations with section/total scores | No additional external sources were available to compare. | |
| | Correlations with external tests | No additional external sources were available to compare. | |
| **Response Processes** | Treatment of atypical | The rubric includes the following instructions: | • Other LLM-specific Concerns: Atypical |

| Type of Evidence | Decision or Analysis to Document | Evidence | Evidence to Address Generative AI-related Concerns |
|---|---|---|---|
| | responses in prompting | *Essays that appear to be off-topic or that pose unusual challenges in other areas should be given a 0.*<br><br>An analysis of advisory codes revealed that both LLMs give much lower scores for responses flagged with excessive length and reuse of language. | Responses (Edge Cases) |
| | Analysis of rationales and/or chain-of-thought results to show consistency with intended use of the rubric | The prompt did not request formal chain-of-thought but did request rationales aimed at helping a human rater understand the score. Rationales were explored qualitatively using some examples to verify that high scores were related to rationales mentioning attributes of high-quality essays, and vice versa. This was also verified by various quantitative analyses of the rationales. We tested if the bigram cosine similarity of pairs of rationales correlate with differences in their associated scores and if the positivity of the sentiment of the rationales is associated with higher scores.<br><br>We did not experiment with alternative prompting strategies that did not request a rationale or chain-of-thought (to determine if rationale/CoT improved score accuracy). | • Lack of Interpretability and Transparency: AI and NLP Explainability Methods<br><br>• Lack of Interpretability and Transparency: Prompting Strategies to Impose Construct Relevant Task Fulfillment |
| | Efforts to minimize and detect prompt injection | No efforts were made to detect prompt injection. | • Other LLM-specific Concerns: Atypical Responses (Edge Cases) |
| | Expert review and/or annotation | There were no expert human reviews of responses and scores conducted for this study. | |
| **Test Content** | Choice of LLM | GPT4, Gemini 1.5 Pro, and Gemini 2.5 Pro were explored for the task. These LLMs are capable of producing evaluations of text by applying a user-provided scoring rubric, but not pretrained to do the specific task of essay scoring.<br><br>• GPT-4 was pre-trained on a broad and diverse dataset containing text from books, websites, academic papers, and more. This included many examples of essays, critiques, and possibly some scored writing samples (OpenAI, 2023).<br><br>• Gemini 1.5 Pro is intended for complex reasoning tasks requiring more intelligence. Gemini 1.5 Pro is a mid-size multimodal model that is optimized for a wide-range of reasoning tasks.<br><br>Gemini 2.5 Pro was trained to be a real-time reasoning model with broader multimodal scope and stronger | • Unknown Training Data: Document LLM Pre-training Data Sources<br><br>• Select LLMs That Were Pretrained to Do the Task |

| Type of Evidence | Decision or Analysis to Document | Evidence | Evidence to Address Generative AI-related Concerns |
|---|---|---|---|
| | | alignment for applied tasks (compared to 1.5 Pro). It was trained on a large-scale, multimodal dataset spanning diverse content types, including web documents, books, and code as well as images, audio, and video. | |
| | Fine-tuning | No fine-tuning was performed. | • No Link to Human Ratings: Fine-tuning |
| | In-context learning (ICL) | No ICL was performed. | • No Link to Human Ratings: In-context learning |
| | Explainability analyses | Rationale analyses support the claim that the scores were based on a process that used the scoring rubric to assign credit. This was verified by various quantitative analyses of the rationales. phrases indicating lower writing quality are associated with lower scores than phrases indicating better writing quality. More generally, embeddings of the rationale's explain 90 percent of the variance in GPT-4o scores. | • Lack of Interpretability and Transparency: AI and NLP Explainability Methods |
| | Reproducibility /Reliability | We used a temperature of 0 for all runs.<br><br>Agreement between scores from two different runs was very high, for both LLMs (GPT-4o and Gemini 1.5 Pro). There were no differences greater than 1 point. | • Lack of Reproducibility/Reliability |
| | Concordance with Human Ratings | Agreement between the LLMs and human ratings were below the typical threshold of QWK = .70 for all models.<br>• GPT-4o and the human ratings: QWK = .52-.53<br>• Gemini 1.5 Pro and the human ratings: QWK = .58<br>• Gemini 2.5 Pro and the human ratings: QWK = .68<br><br>The LLM agreement rates were much lower compared to the e-rater model (QWK = .84) | |
| | Quality of Evaluation Sample | According to Crossley et al. (2024), raters were trained for each prompt and essay type. The responses were scored by human raters using a 100% adjudication approach in which each response was scored by two raters and any responses with any score differences were adjudicated by a third rater. Therefore, the PERSUADE dataset contained 1 human rating. Before the adjudication process, the weighted kappa was .745. | |
| | Concordance with e-rater | The QWK between each LLM and e-rater was similar to their QWKs with the human rating: GPT-4o: .50, Gemini 1.5 Pro: .57, Gemini 2.5 Pro: .65) | |

| Type of Evidence | Decision or Analysis to Document | Evidence | Evidence to Address Generative AI-related Concerns |
|---|---|---|---|
| | Correlations between scores and NLP features | Correlations between the LLM scores and the e-rater feature values showed that:<br><br>• The LLMs have positive correlations with all the features except *Discourse Coherence*, which also has weak or negative correlation with human ratings and the other features.<br>• The correlations between LLMs and e-rater features are similar to the corresponding correlations for human ratings or e-rater and the features except the correlations between LLMs and *Usage* are higher than those for human ratings or e-rater.<br>• The correlations for *Development*, *Organization*, and *Sentence Variety* tend to be smaller for the LLM scores than human ratings and e-rater. The LLMs are not as sensitive to essay length as the humans and e-rater. | |
| | Expert annotations | No human expert annotations were available. | |
| | Inter-item correlations and correlations between item and section scores. | No additional external sources were available to compare. | |
| Consequences of Use | Analysis of unintended and intended consequences | Since this was an exploratory study, no data were available to study the consequences of score use. | |
| Fairness | Subgroup analyses comparing results based on human ratings vs AI, by subgroup | • Most SMDs across groups and models were close to zero ($\leq 0.05$), showing no consistent bias.<br><br>• Exception: GPT-4o and Gemini 1.5 Pro showed SMDs around −0.20 for Asian/Pacific Islander students, indicating systematically lower scores than human raters. | • Unknown Training Data: Design a System to Detect, Understand, and Mitigate Bias |
| | Fairness of human ratings used in fine-tuning/ICL | No fine-tuning or ICL was used. | |
| | Review of AI explainability analyses for unfairness | Analyzed e-rater features by race/ethnic group. Discovered that the Asian/Pacific Islander group had high quality essays on average, and wrote longer essays. GPT-4o and Gemini 1.5 Pro possibly underweighted essay length compared to human raters. This is a potential source of the large SMDs. | • Unknown Training Data: Design a System to Detect, Understand, and Mitigate Bias<br>• Lack of Interpretability and Transparency: AI |

| Type of Evidence | Decision or Analysis to Document | Evidence | Evidence to Address Generative AI-related Concerns |
|---|---|---|---|
| | | | and NLP Explainability Methods |

## 5. Discussion

The purposes of this paper were to highlight the differences in the feature-based and generative AI applications in CR scoring systems and discuss how validity argumentation should also differ as a result. The set of validity evidence for generative AI-based scores will be different and possibly more extensive. The main differences in the necessary evidence relate to the unknown training data for the LLMs, the lack of reproducibility, the lack of transparency of LLMs and perhaps most importantly, the lack of link to human ratings. We explore those differences by outlining studies, analyses and specific sources of evidence that would contribute to overcoming those limitations. We also provide an example of how the collection of evidence might look for the application of GPT-4o and Gemini 1.5 Pro and 2.5 Pro to CR scoring of essays from the PERSUADE corpora.

We wrote this manuscript to aid AI scientists and engineers using generative AI to score tests. While this group has their own principles learned from data science, machine learning, NLP, and other technical areas, they may not have the expertise in psychometric principles and validity theory. We believe that in these scenarios, cross-functional teams must work together to ensure that the results were a product of decisions made in a principled fashion using the validity framework as specified in our industry standards (AERA, APA, & NCME, 2014) and as proposed in this paper.

Though not nearly as transparent and connected to the construct compared to the traditional NLP version of automated scoring, there are several opportunities for a human-in-the-loop to positively influence the results of the LLM predictions. The engineer is often the human-in-the-loop involved in the first stages of the experiments for LLM selection, fine-tuning, etc. We invite engineers to consult with psychometricians and subject matter experts at these beginning stages to help with decision-making, as appropriate. Importantly, while LLMs may offer faster and cheaper scoring than human raters or traditional feature-based AI scoring models, there is potentially a large expense in performing due diligence studies and curating validity evidence. It is important to pursue evidence beyond quantitative comparisons; we encourage scientists to consider qualitative analyses including think aloud protocols conducted by subject matter experts, reviews of LLM-generated rationales, and review of atypical responses and how the LLM scores them. Using humans for ongoing review can help in evaluating and improving the LLM pipeline. For example, performing a comprehensive evaluation of the scores and their meaning involves a review of a sample of responses by subject matter experts, reviewing saliency results, and more. This type of work is time-consuming and costly, but if it is not done, then there may be insufficient evidence to support the use and interpretation of the scores. For this reason, using an off-the-shelf LLM for CR scoring may be less cost-effective than expected or hoped.

### 5.1 The Diminishing Human-in-the-Loop?

The natural next step after integrating generative AI-based scores in CR scoring systems is the version that minimizes or completely excludes human oversight. Would it be possible to imagine a CR scoring system with no expert-defined rubric? No annotations? No established concordance with human ratings? No human monitoring post-deployment? The only human-in-the-loop might be the engineer. Because CRs are an important part of assessment, their use will continue and may be expanded in the future of assessment. However, the future of CR scoring will likely look very different – perhaps a construct definition is all that is needed for item and rubric development by AI. As we move toward more AI and more automation, we need to determine the minimum amount of human involvement needed for validity evidence, especially in the high-stakes context. Can there be sufficient evidence without any comparison of human scores to human judgments of the response? If so, what other evidence would allow for full automation to be acceptable? As the capabilities of AI continue to evolve, standards for the validity of scores will also need to evolve.

**Appendix A**

```
You are a seasoned English teacher who has been trained to be a professional
rater of high school essays. Use the instructions in the provided rubric to
evaluate and score the response to the assigned question.

First start with reading the response text and then consider how the response
aligns with the rubric indicators. Note that the rubric details a holistic
scoring process. Don't be too analytical or detailed when assessing the
responses. We want an overall score.

In addition to the score, please also provide your reasoning or your thought
process for the score you assigned. Please describe it in no more than 2
sentences so that another rater could use it to understand your score.

The question or task, rubric, and answer will each be surrounded with XML-style
tags below. The tags will be\n\n",
    "<D5A60FF8F3AF47619BC1CE00CA21D938></D5A60FF8F3AF47619BC1CE00CA21D938>,\n",
    "<27152C7AC19445FA87D5FC4A7313FF68></27152C7AC19445FA87D5FC4A7313FF68>,
and\n",
    "<CACE4B6E785148BDAD20A93818F662B8></CACE4B6E785148BDAD20A93818F662B8>,
respectively. \n\n",

Return your answer in this exact JSON format:
    "{\n  \"score\": <number>,\n  \"reasoning\": \"<brief explanation>\"\n}
```

**Fig 5** Prompt used in demonstrative study to request essay scores from LLMs.

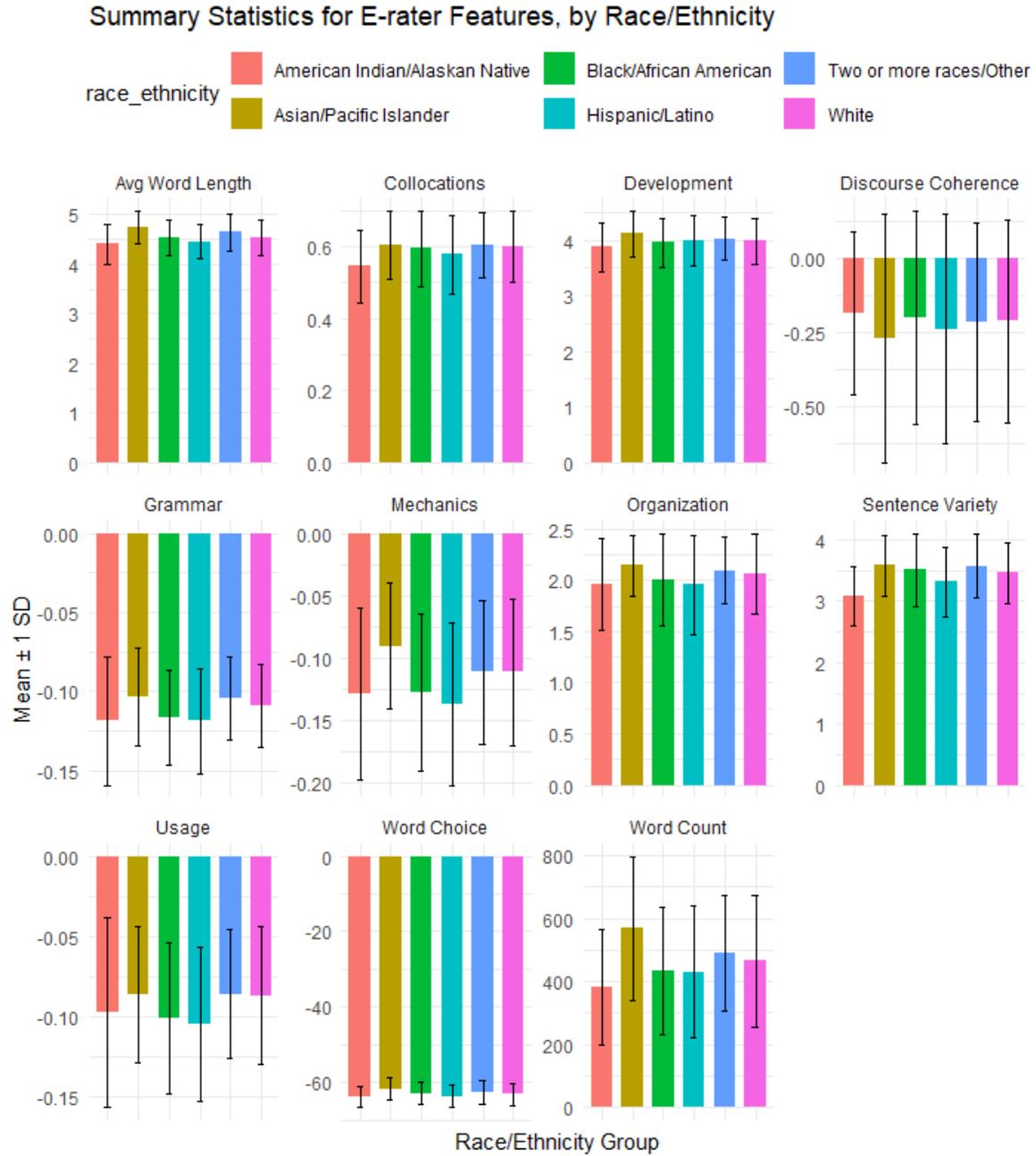

**Fig 6** Mean of e-rater features (raw) and essay length, for each race/ethnicity group.

**Table 7** Distribution of AI (machine) scores conditional on human rating for PERSUADE 2.0 data. Shows the row percentages for each AI model at each human rating level.

| Human | | AI Model | | | | | |
|---|---|---|---|---|---|---|---|
| | | **1** | **2** | **3** | **4** | **5** | **6** |
| **1**<br>**(n = 42)** | **E-rater** | 19 | 62 | 14 | 5 | 0 | 0 |
| | **GPT-4o** | 31 | 67 | 2 | 0 | 0 | 0 |
| | **Gemini 1.5 Pro** | 26 | 69 | 5 | 0 | 0 | 0 |
| | **Gemini 2.5 Pro** | 36 | 64 | 0 | 0 | 0 | 0 |
| **2**<br>**(n = 1,018)** | **E-rater** | 4 | 61 | 31 | 3 | 0 | 0 |
| | **GPT-4o** | 2 | 67 | 30 | 0 | 0 | 0 |
| | **Gemini 1.5 Pro** | 1 | 63 | 32 | 4 | 0 | 0 |
| | **Gemini 2.5 Pro** | 2 | 56 | 41 | 0 | 0 | 0 |
| **3**<br>**(n = 1,862)** | **E-rater** | 0 | 10 | 66 | 23 | 1 | 0 |
| | **GPT-4o** | 0 | 35 | 60 | 5 | 0 | 0 |
| | **Gemini 1.5 Pro** | 0 | 25 | 53 | 22 | 0 | 0 |
| | **Gemini 2.5 Pro** | 0 | 15 | 73 | 11 | 0 | 0 |
| **4**<br>**(n = 2,009)** | **E-rater** | 0 | 0 | 14 | 75 | 11 | 0 |
| | **GPT-4o** | 0 | 11 | 60 | 28 | 0 | 0 |
| | **Gemini 1.5 Pro** | 0 | 7 | 34 | 58 | 1 | 0 |
| | **Gemini 2.5 Pro** | 0 | 2 | 49 | 43 | 6 | 0 |
| **5**<br>**(n = 1,210)** | **E-rater** | 0 | 0 | 0 | 28 | 69 | 2 |
| | **GPT-4o** | 0 | 1 | 36 | 56 | 6 | 0 |
| | **Gemini 1.5 Pro** | 0 | 0 | 11 | 85 | 3 | 0 |
| | **Gemini 2.5 Pro** | 0 | 0 | 17 | 53 | 26 | 4 |
| **6**<br>**(n = 364)** | **E-rater** | 0 | 0 | 0 | 1 | 70 | 29 |
| | **GPT-4o** | 0 | 0 | 12 | 60 | 27 | 0 |
| | **Gemini 1.5 Pro** | 0 | 0 | 3 | 82 | 13 | 1 |
| | **Gemini 2.5 Pro** | 0 | 0 | 4 | 28 | 49 | 19 |

**Table 8  E-rater engine advisories with mean and standard deviations of scores**

| Advisory | n | Human | | E-rater | | GPT-4o | | Gemini 1.5 Pro | |
|---|---|---|---|---|---|---|---|---|---|
| | | *M* | *SD* | *M* | *SD* | *M* | *SD* | *M* | *SD* |
| No Advisories | 5,125 | 3.51 | 1.06 | 3.50 | 0.93 | 2.99 | 0.75 | 3.23 | 0.80 |
| Reuse of Language | 829 | 3.87 | 0.92 | 3.97 | 0.65 | 3.02 | 0.76 | 3.37 | 0.80 |
| Excessive Length | 298 | 5.51 | 0.63 | 5.35 | 0.34 | 4.00 | 0.70 | 4.04 | 0.52 |
| Reuse of Language + Excessive Length | 166 | 4.99 | 0.88 | 5.07 | 0.35 | 3.40 | 0.81 | 3.69 | 0.69 |
| Unidentifiable Organizational Elements | 53 | 2.58 | 0.75 | 3.01 | 0.68 | 2.32 | 0.61 | 2.45 | 0.72 |
| Reuse of Language + Unidentifiable Organizational Elements | 11 | 2.64 | 1.03 | 3.25 | 0.78 | 2.27 | 0.91 | 2.55 | 0.93 |
| Reuse of Language + Excessive Length + Unidentifiable Organizational Elements | 8 | 4.12 | 1.36 | 4.63 | 0.51 | 2.62 | 0.92 | 3.12 | 0.99 |
| Too Brief + Unidentifiable Organizational Elements | 7 | 1.71 | 0.49 | 1.41 | 0.26 | 2.00 | 0.00 | 1.86 | 0.38 |
| Excessive Length + Unidentifiable Organizational Elements | 7 | 4.43 | 1.40 | 4.97 | 0.62 | 3.00 | 0.82 | 3.43 | 0.98 |
| Reuse of Language + Too Brief + Unidentifiable Organizational Elements + Excessive Number of Problems | 1 | 2.00 | -- | 2.85 | -- | 1.00 | -- | 2.00 | -- |
| **All Responses** | **6,505** | **3.68** | **1.13** | **3.68** | **0.99** | **3.04** | **0.79** | **3.29** | **0.81** |